\documentclass[sigconf]{acmart}

\usepackage{courier} 
\usepackage{graphicx} 
\usepackage{caption} 
\usepackage{amsmath}
\usepackage{amsfonts}

\usepackage{amssymb}
\usepackage{amsthm}
\usepackage{multirow}
\usepackage{makecell}
\usepackage{booktabs}
\usepackage{xcolor}
\usepackage{enumitem}
\usepackage{verbatim}
\usepackage[ruled]{algorithm2e}
\usepackage{subfigure} 
\usepackage{ragged2e}
\usepackage{caption}
\usepackage{appendix}
\usepackage{bm}
\usepackage{longtable}
\usepackage{adjustbox}
\usepackage{microtype}
\usepackage{setspace}
\usepackage{hyperref}
\usepackage{float}
\usepackage{balance}

\usepackage{nicefrac}      
\usepackage{wrapfig}
\usepackage{pifont}
\usepackage{threeparttable}

\theoremstyle{plain}
\newtheorem{theorem}{Theorem}[section]
\newtheorem{proposition}[theorem]{Proposition}
\newtheorem{lemma}[theorem]{Lemma}
\newtheorem{corollary}[theorem]{Corollary}
\theoremstyle{definition}
\newtheorem{definition}[theorem]{Definition}

\theoremstyle{remark}
\newtheorem{remark}[theorem]{Remark}

\AtBeginDocument{%
  }

\setcopyright{acmlicensed}
\copyrightyear{2026}
\acmYear{2026}

\acmConference[KDD '26]{Proceedings of the 32st ACM SIGKDD Conference on Knowledge Discovery and Data Mining V.1}{August 9--13, 2026}{Jeju, ON, South Korea}
\acmDOI{XXXXXXX.XXXXXXX}
\acmISBN{978-1-4503-XXXX-X/2018/06}

\begin{document}


\title{Spatiotemporal Graph Learning with Direct Volumetric Information Passing and Feature Enhancement}


\author{Yuan Mi}
\affiliation{%
  \institution{Renmin University of China}
  \city{Beijing}
  \country{China}}
\email{miyuan@ruc.edu.cn}

\author{Qi Wang}
\affiliation{%
  \institution{Renmin University of China}
  \city{Beijing}
  \country{China}}
\email{qi_wang@ruc.edu.cn}

\author{Xueqin Hu}
\affiliation{%
  \institution{Wuhan University of Technology}
  \city{Wuhan}
  \country{China}}
\email{hxq@whut.edu.cn}

\author{Yike Guo}
\affiliation{%
  \institution{Hong Kong University Of Science and Technology}
  \city{Hong Kong}
  \country{China}}
\email{yikeguo@ust.hk}

\author{Ji-Rong Wen}
\affiliation{%
  \institution{Renmin University of China}
  \city{Beijing}
  \country{China}}
\email{jrwen@ruc.edu.cn}

\author{Yang Liu}
\affiliation{%
  \institution{University of Chinese Academy of Sciences}
  \city{Beijing}
  \country{China}}
\email{liuyang22@ucas.ac.cn}

\author{Hao Sun}
\authornote{Corresponding authors.}
\affiliation{%
  \institution{Renmin University of China}
  \city{Beijing}
  \country{China}}
\email{haosun@ruc.edu.cn}

\renewcommand{\shortauthors}{Yuan Mi et al.}

\begin{abstract}

Data-driven learning of physical systems has kindled significant attention, where many neural models have been developed. In particular, mesh-based graph neural networks (GNNs) have demonstrated significant potential in modeling spatiotemporal dynamics across arbitrary geometric domains. However, the existing node-edge message-passing and aggregation mechanism in GNNs limits the representation learning ability. In this paper, we proposed a dual-module framework, Cell-embedded and Feature-enhanced Graph Neural Network (aka, CeFeGNN), for learning spatiotemporal dynamics. Specifically, we embed learnable cell attributions to the common node-edge message passing process, which better captures the spatial dependency of regional features. Such a strategy essentially upgrades the local aggregation scheme from first order (e.g., from edge to node) to a higher order (e.g., from volume and edge to node), which takes advantage of volumetric information in message passing. Meanwhile, a novel feature-enhanced block is designed to further improve the model's performance and alleviate the over-smoothness problem. Extensive experiments on various PDE systems and one real-world dataset demonstrate that CeFeGNN achieves superior performance compared with other baselines. 
\end{abstract}




\maketitle

\vspace{-2pt}
\section{Introduction}\label{Introduction}
Neural surrogate models are often essential for analyzing and modeling complex spatiotemporal dynamics processes across various scientific and engineering fields. For example, weather prediction ~\cite{scher2018toward,schultz2021can,grover2015deep}, ocean current motion prediction ~\cite{zheng2020purely}, nonlinear engineering structure earthquake response prediction ~\cite{zhang2019deep}, material mechanical properties simulation ~\cite{wang2018multiscale}, etc. Traditionally, classical numerical methods (e.g., Finite Difference Method (FDM) ~\cite{godunov1959finite,ozicsik2017finite}, Finite Volume Method (FVM) ~\cite{eymard2000finite}, and Finite Element Method (FEM)~\cite{hughes2012finite}) are utilized to solve these PDEs, requiring substantial analytical or computational efforts. Although this problem has been simplified via discretizing the space, the issue of trade-off between cost and precision intensifies when dealing with varying domain geometries (e.g., different initial or boundary conditions or various input parameters), especially in real-world scenarios. In the last few decades, Deep Learning (DL) ~\cite{pinkus1999approximation,tolstikhin2021mlp,albawi2017understanding,koutnik2014clockwork,sundermeyer2012lstm}  models have made great progress in approximating high-dimensional PDEs benefiting from existing rich labeled or unlabeled datasets. However, there are certain drawbacks in this simple approach of learning the non-linear mapping between inputs and outputs from data. For example, their performance is severely limited by the training datasets, the neural network lacks interpretability and generalizes poorly.

\begin{figure*}[t!]
	\centering
	\includegraphics[width=0.98\textwidth]{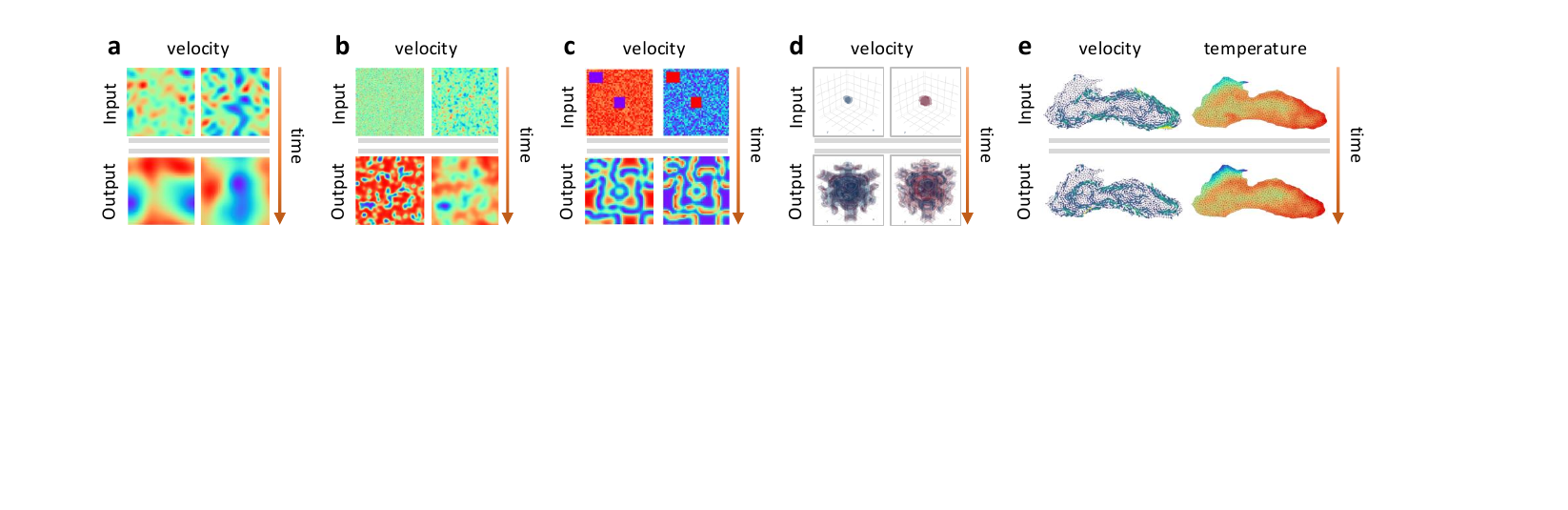}
	\vspace{-6pt}
	\caption{Examples of datasets, including classic governing equations and complex real-world dataset.
		\textbf{a}, the 2D Burgers equation.
		\textbf{b}, the 2D Fitzhugh-Nagumo equation.
		\textbf{c}, the 2D Gray-Scott equation.
		\textbf{d}, the 3D Gray-Scott equation.
		\textbf{e}, the 2D Black-Sea dataset.
	}
	\label{fig:examples}
	\vspace{-6pt}
\end{figure*}

Embedding domain-specific expertise (e.g., Physics-informed Neural Networks (PINNs) ~\cite{raissi2019physics}) has shown the potential to tackle these problems ~\cite{krishnapriyan2021characterizing,gao2021phygeonet,he2023learning,li2024physics}. However, the core part of PINNs, \textit{Automatic Differentiation} (AD) approach, has two major drawbacks: (1) it is necessary to formulate explicit governing equations into the loss function, and (2) the parameters in high-dimensional feature spaces cannot be efficiently optimized when facing highly complex networks like graph networks. As shown in Figure \ref{fig:examples}\textbf{e}, there are no any predefined equations available to represent the evolution patterns of sea surface temperature at varying depths. Neural Operators, such as DeepONet ~\cite{lu2021learning} and Latent DeepONet ~\cite{kontolati2024learning}, have emerged as another paradigm to learn these complex non-linear behaviors. The most well-known models, Fourier Neural Operator (FNO) ~\cite{li2021Fourier} and its variants ~\cite{tran2021factorized,wen2022u,ashiqur2023u,li2024geometry}, utilize neural networks to learn parameters in the Fourier space for fast and effective turbulence simulation. Likewise, they inevitably have the same shortcoming as those traditional methods: over-reliance on data and biased towards the grid domain.

The other representative models, Transformer models ~\cite{vaswani2017attention,wu2024transolver} and GNNs ~\cite{liu2020towards,gao2022physics,mccardle2023shedding,horie2024graph}, have demonstrated significant influence in predicting spatiotemporal dynamics across arbitrary geometric domains. In particular, mesh-based graph neural networks (GNNs) ~\cite{gilmer2017neural,gilmer2020message,pfaff2021Learning,brandstetter2022message} learn vastly different dynamics of various physical systems, ranging from structural mechanics and cloth to fluid simulations. However, the existing node-edge message passing mechanism in GNNs overestimates the primary role of ``message'' passing function on the neighbor ``edges'', limiting the model's representation learning ability. In general, this strategy leads to highly homogeneous node features after multiple rounds of message passing, making the features ineffective at representing distinct characteristics, namely, the over-smoothness problem.

\begin{figure*}[t!]
\centering
\includegraphics[width=0.95\textwidth]{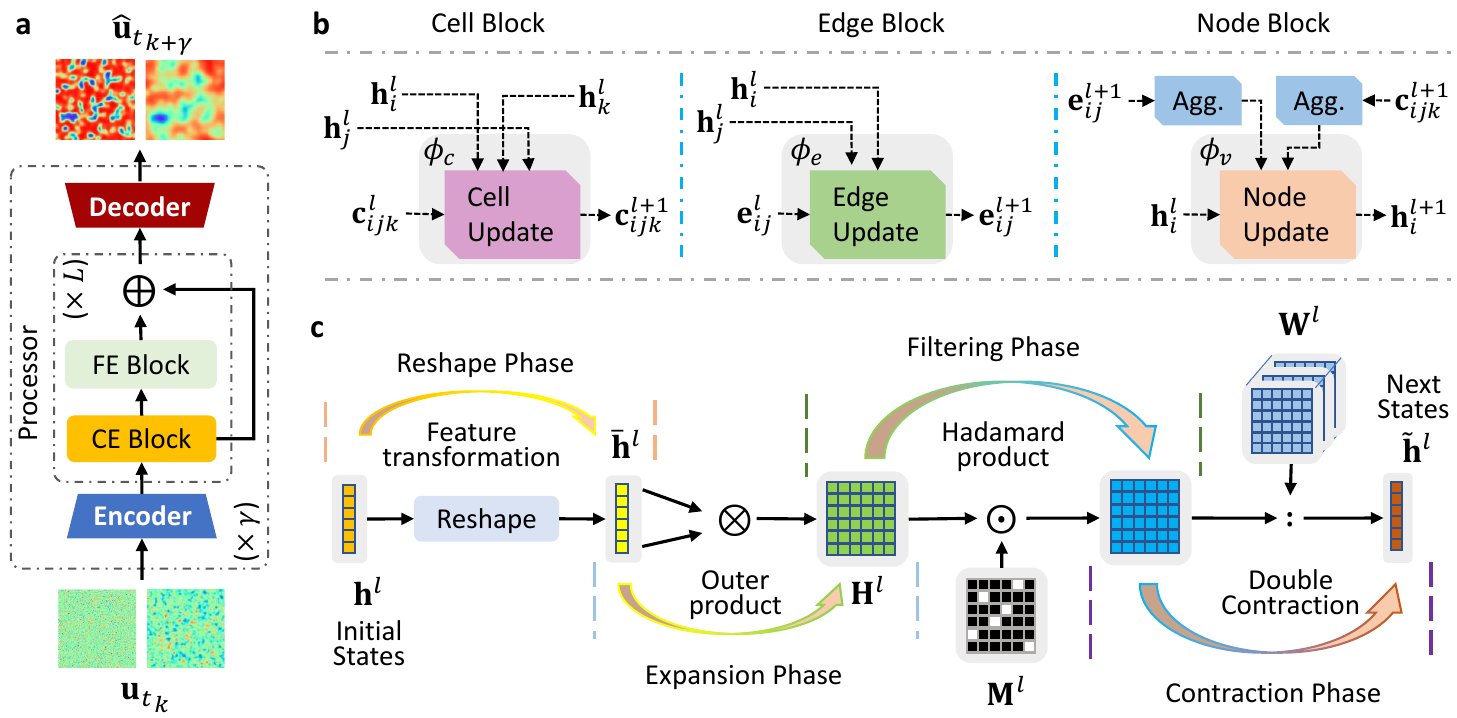}
\vspace{-6pt}
\caption{Network architecture of CeFeGNN. \textbf{a}, an encoder encodes the physical variables to latent features, multiple message passing blocks process these latent features iteratively, and a decoder maps back to the physical states. \textbf{b}, three components in the CE block. \textbf{c}, the process of FE block. The definition of relative symbols refers to Section \ref{Methodology}.}
\label{fig:architecture}
\vspace{-6pt}
\end{figure*}

To further address the above issues, we proposed an end-to-end graph-based framework, Cell-embedded and Feature-enhanced Graph Neural Network (CeFeGNN), to model spatiotemporal dynamics across various domains with improved performance. 
Specifically, after detecting discontinuities in space, we embed learnable cell attributions into the existing node-edge message passing process, which better captures the spatial dependency of regional features and legitimately learns higher-order information from connected nodes of any cell. This cell-embedded (CE) method allows us to rapidly identify non-local relationships that traditional message-passing mechanisms often fail to capture directly. Essentially, such a strategy upgrades the local aggregation scheme from the first order (e.g., from edge to node) to a higher order (e.g., from volume and edge to node), which takes advantage of volumetric information in message passing. Meanwhile, a novel feature-enhanced (FE) block is designed to further improve the model's performance and relieve the over-smoothness problem via treating the latent features as basis functions and further processing these features on this concept. In details, it regards the node feature $\mathbf{h}_{i}$ as basis and builds a higher-order feature via an outer-product operation, e.g., $\mathbf{h}_{i}\otimes\mathbf{h}_{i}$. This process creates abundant high-order nonlinear terms to enrich the feature map when it is applied iteratively. We then use a mask operation to randomly sample these terms, filtering the appropriate information by a learnable weight tensor to enhance the model's representation capacity. Figure \ref{fig:architecture} shows an outline of our proposed model. Extensive experimental results on many PDE-centric systems and real-world datasets show that CeFeGNN significantly enhance spatiotemporal learning in various scenarios, particularly with limited training datasets. 
The dual-module structure based on the synergy between the cell-embedded module and the FE module enables our model to effectively extract and process interaction features that are better serving for for complex dynamics learning. The key contributions are summarized as follows:
\begin{itemize}
    \item We introduce a two-level cell-embedded message-passing mechanism to better capture the spatial dependency on complex domains, which is not only easy to follow but also lightweight yet effective.
    \item We propose a unique feature-enhanced (FE) block to learn the high-order features and capture dynamic interactions, which conforms to empirical intuition. 
    \item Extensive experiments demonstrate that our approach stands out for its lower error and robust generalizability in the spatiotemporal dynamic field.
\end{itemize}

\vspace{-4pt}

\vspace{-2pt}
\section{Related Works}\label{Related Works}
Spatiotemporal dynamics research, as one of the important frontier research areas, is integral to fields ranging from traditional fluid dynamics to economics and finance. In this part, we firstly give a brief introduction to spatiotemporal PDEs. Then, the relevant progress in spatiotemporal dynamics research is described from the perspectives of classical and neural solvers. 

\vspace{-2pt}
\subsection{Problem statement}\label{Problem formulation}
Without loss of generality, the time-dependent systems generally describe the evolution of a continuous field over certain time intervals, which can be cast into the following form:
$\partial \mathbf{u} / \partial t={\mathcal{F}} \left  (\theta, t, \mathbf{x}, \mathbf{u}, \boldsymbol{\nabla} \mathbf{u}, \boldsymbol{\nabla}^{2} \mathbf{u}, \dots \right ),$
where $\mathcal{F} (\cdot)$ denotes an unknown linear or nonlinear function comprised of the spatiotemporal variable $\mathbf{u} (\mathbf{x}, t) \in\mathbb{R}^{{d}}$, its corresponding partial derivatives  (e.g., $\boldsymbol{\nabla} \mathbf{u}, \boldsymbol{\nabla}^{2} \mathbf{u}$), and some related parameters $\theta$. Here, $\mathbf{x}\in\mathbb{R}^m$ denotes the $m$-dimensional spatial coordinate, $t \in \mathbb{R}^{1}$ the time, $\boldsymbol{\nabla}$ the Nabla operator, $\boldsymbol{\nabla}^{2}$ the Laplacian operator. Our task is \textit{\textbf{to predict the states of all subsequent moments after a random initial condition}}.

\vspace{-2pt}
\subsection{Classical solvers}
To solve time-dependent PDEs, a common way is the method of lines (MOL). By discretizing in all but one dimension, it allows solutions to be computed via methods and software developed for the numerical integration of ordinary differential equations (ODEs) and differential-algebraic equations (DAEs). Meanwhile, the multigrid method \cite{wu2020multigrid} is another algorithm for solving PDEs via a hierarchy of discretizations. Other classical numerical methods (e.g., Finite Difference Method (FDM) \cite{godunov1959finite,ozicsik2017finite}, Finite Volume Method (FVM)\cite{eymard2000finite}, and Finite Element Method (FEM)\cite{cao1999anr}) have also been utilized for practical applications \cite{reich2000finite,hughes2012finite}. 

\vspace{-2pt}
\subsection{Neural solvers}

\subsubsection{PINN Methods}
Two main approaches, Physics-informed Neural networks (PINNs) \cite{raissi2019physics,krishnapriyan2021characterizing,he2023learning} and Physics-informed Neural Operators (PINOs) \cite{li2021physics,hao2023gnot,kovachki2023neural}, {were developed} to learn fluid and solid mechanisms. With formulating the explicit governing equation as the loss function, PINNs and PINOs constrain the latent feature spaces to a certain range, effectively learning from small data or even without any labeled data. Such novel methods immediately attract the attention of many researchers and has been utilized in a wide range of applications governed by differential equations, such as heat transfer problems \cite{cai2021physics}, power systems \cite{misyris2020physics}, medical science \cite{sahli2020physics}, and control of dynamical systems \cite{antonelo2024physics}. 

\vspace{-2pt}
\subsubsection{Neural Operators}
Common neural operators \cite{lu2021learning,kontolati2024learning} combine various basis transforms (e.g., Fourier, multipole kernel, wavelet \cite{gupta2021multiwavelet}) with neural networks to accelerate PDE solvers in diverse applications. For example,  Fourier Neural Operator (FNO) and its variants \cite{tran2021factorized,wen2022u,ashiqur2023u,li2024physics} learn parameters in the Fourier domain for turbulence simulation. Especially, Geo-FNO \cite{li2024geometry} maps the irregular domain into an uniform mesh with a specific geometric Fourier transform to fit irregular domains. However, they all follow the assumptions of periodicity and time-invariance property, making them fail in complex boundaries. Their excellent performance ``illusion'' in turbulent flow is primarily due to their turbulence solver closely related to the Fourier transform.

\vspace{-2pt}
\subsubsection{Transformer Methods}
As another paradigm, Transformer \cite{vaswani2017attention} and its ``x-former'' family  \cite{jiang2023pdformer,wu2024transolver} have also been utilized to solve complex PDEs. Given the attention mechanism along with a higher complexity, many researchers are trying to alleviate this issue through various means. For example, linear attention mechanism \cite{li2022transformer,hao2023gnot} is a well-known method to address this limitation. Although the above methods alleviate the need for specific domain expertise, they all share the same limitations: instability in long-range prediction and weak generalization ability.

\vspace{-2pt}
\subsubsection{Graph Methods}
Abundant works about Graph Neural Networks (GNNs) \cite{liu2020towards,gao2021phygeonet,gao2022physics,horie2024graph} and geometric learning \cite{bronstein2017geometric,hajij2020cell,hajij2022topological,horie2024graph} attempt to utilize customized substructures to generalize message passing to more complex domains. For example, graph kernel methods \cite{anandkumar2020neural,li2020multipole} try to learn the implicit or explicit embedding in Reproducing Kernel Hilbert Spaces (RKHS) for identifying differential equations. \cite{belbute2020combining} considers solving the problem of predicting fluid flow using GNNs. \cite{lienen2022learning} introduces the interpolation strategy and the finite element method into GNNs. \cite{fortunato2022multiscale} constructs the multi-scale message-passing dependency. \cite{bodnar2021weisfeiler} perform message passing on simplicial complexes (SCs). Most notably, MPNNs \cite{gilmer2017neural,gilmer2020message,janny2023eagle,perera2024multiscale} are utilized to tackle these issues, which learn latent representations on graphs via message passing mechanism. Especially, MeshGraphNets \cite{pfaff2021Learning} and MP-PDE solver \cite{brandstetter2022message} are two representative examples. While powerful and expressive, we still find that there are some problems with these methods: \textbf{\textit{(1) their expression ability is still not strong enough, (2) their prediction error is still somewhat high, and (3) they still rely on a large amount of data}}. Hence, we designed our model to address these three problems.

\vspace{-2pt}
\section{Methodology}\label{Methodology}

\begin{figure*}[t!]
	\centering
	\includegraphics[width=0.95\textwidth]{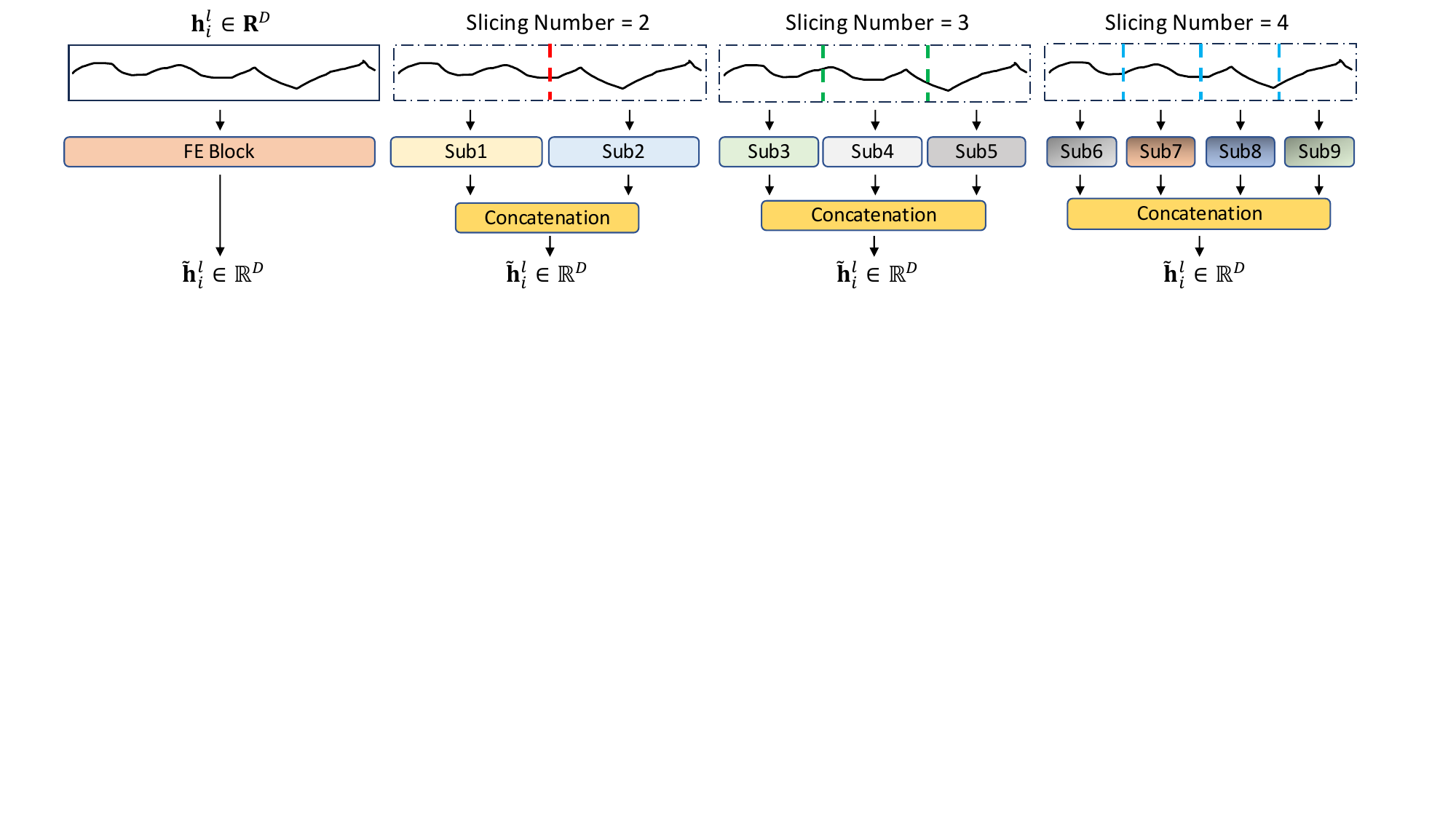}
	\vspace{-6pt}
	\caption{A scheme for reducing the number of parameters in FE block. Its quantitative experiment of the impact of window size and number of sub-features on CeFeGNN is shown in Table \ref{tab:impact_of_window_size_and_number_of_sub-features}.}
	\label{fig:fe}
	\vspace{-6pt}
\end{figure*}

In this section, we illustrate how our method effectively learn the solution of spatiotemporal PDEs under various parameters (e.g., the initial or boundary conditions, constant or variable coefficients) {for} a given physical system. All the source code and data would be posted after peer review.

\subsection{Network architecture}\label{Network architecture}
To enhance the performance of long-range prediction, we adapt the conventional ``Encoder-Processor-Decoder'' framework in ~\cite{pfaff2021Learning,brandstetter2022message} as the backbone of our method, which is primarily designed to effectively learn the complex spatiotemporal dependencies on graphs. As shown in Figure \ref{fig:architecture}, our proposed method mainly consists of the Feature-Enhanced (FE) block and the Cell-Embedded (CE) block. 
These two key components update features in a sequential process to achieve the cascaded enhancement effect that: (1) Updating node-edge-cell features with the CE block; (2) Enriching higher-order node features with the FE block; (3) Iteratively repeating the above two steps until the specified number of processor layers is reached. The synergy of these two sequentially placed blocks in turn improves the model's representation learning capacity and generalization ability. The source code and data are provided at \url{https://www.github.com} (posted after the peer-review process).

\subsubsection{Feature-enhanced block}\label{subsubsec:fe}
For the sake of brevity and clarity, FE block, shown in Figure \ref{fig:architecture}, is designed to enhance the latent features from the upstream block and further alleviate the over-smoothness issue commonly seen in GNNs due to excessive aggregation. Please see the ablation results on FE's efficacy in Table \ref{tab:Ablation Study}. More details are provided in Appendix Section \ref{Mathematical Inspiration of Feature-enhanced (FE) Block}.

\paragraph{\textbf{Outer Product as Basis Expansion.}}
The outer product operation $\otimes$ on the reshaped feature map $\overline{ \mathbf{h}}_{i} \in \mathbb{R}^{D \times 1}$ expands the original latent feature space into a higher-order tensor space. This expansion introduces second-order terms (e.g., $\alpha\beta$ for $\{\alpha, \beta\}\in\overline{ \mathbf{h}}_{i}$), which can capture interactions between individual components of the original feature $\mathbf{h}_{i} \in \mathbb{R}^{D}$. 
Mathematically, the second-order tensor reads $\overline{ \mathbf{h}}_{i} \otimes \overline{ \mathbf{h}}_{i}$. This operation creates a richer feature map with cross-term interactions that may not be explicitly encoded in the original latent space.

\begin{lemma}[Nonlinear Representation]
The second-order terms $\alpha\beta$ can model nonlinear dependencies between features. This is particularly useful for capturing complex interactions that linear transformations (e.g., via simple dot products) might overlook.
\end{lemma}

\begin{definition}
The FE block expands the latent feature $\overline{\mathbf{h}}_{i} \in \mathbb{R}^{D\times 1}$ of node $i$ into a higher-order feature map $ \mathbf{H}_{i} \in \mathbb{R}^{D\times D} $ using an outer product: $ \mathbf{H}_{i} = \overline{\mathbf{h}}_{i} \otimes \overline{\mathbf{h}}_{i}$.
\end{definition}

\paragraph{\textbf{Regularization via Masking.}}
Masking introduces sparsity in $\mathbf{H}_{i}$, reducing overfitting. If $\mathbf{M}_{jk}$ is selected, $\mathbf{M}_{jk} = 1$. Otherwise, $\mathbf{M}_{jk} = 0$. Here $j$ and $k$ index the $M \in \mathbb{R}^{D\times D}$ components.


\paragraph{\textbf{Learnable Filtering.}}
The learnable weight tensor $\mathbf{W} \in \mathbb{R}^{D\times D\times D}$ acts as a filter, selecting and emphasizing the most informative terms from upstream tasks. 

\begin{definition}
A mask tensor $\mathbf{M} \in \mathbb{R}^{D\times D}$ is applied to randomly sample elements in $\mathbf{H}_{i}$, and the resulting masked tensor is processed using a learnable weight tensor $\mathbf{W} \in \mathbb{R}^{D\times D\times D}$ as follows: $\tilde{\mathbf{h}}_{i} = ( \mathbf{M} \odot\mathbf{H}_{i}) : \mathbf{W}, $ where $\odot$ represents element-wise multiplication and $:$ denotes double contraction of tensors. The resulting feature $\tilde{\mathbf{h}}_{i} \in \mathbb{R}^{1\times D}$ enriches the representation of $\mathbf{h}_{i} \in \mathbb{R}^{D}$.
\end{definition}

\begin{corollary}[Representation Power]
The full feature map $\mathbf{H}_{i}$ contains $D^2$ terms for a $D$-dimensional input feature vector $\mathbf{h}_{i}$. After masking, the effective representation space reduces by the sparsity of $\mathbf{M}$. The learnable filter $\mathbf{W}$ further narrows this down to the most critical terms.
\end{corollary}

\begin{figure}[t!]
\centering
\includegraphics[width=0.65\columnwidth]{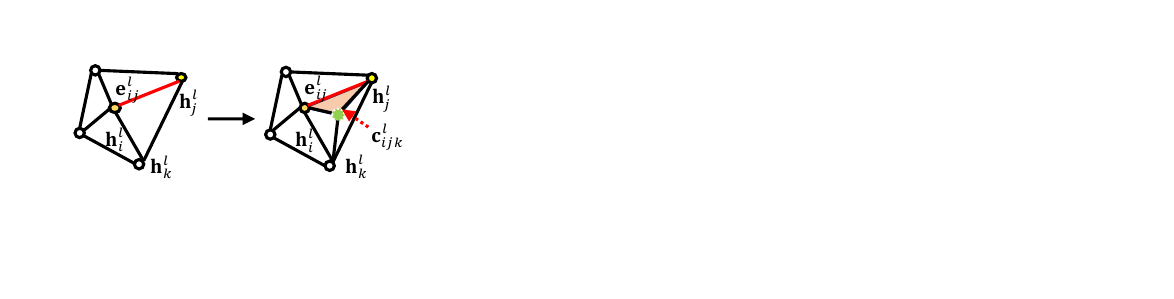}
\vspace{-6pt}
\caption{Cell in graphs. Green point is the centroid of cell. 
}
\label{fig:cell}
\vspace{-12pt}
\end{figure}

Considering the GPU memory requirements caused by additional parameters in the FE module, we further provide a feature splitting scheme within latent features to dramatically reduce the number of parameters and computation cost caused by the full FE block (see Figure \ref{fig:fe}). This strategy divides every feature into multiple sub-features with different window sizes, and then processes each part separately. Finally, these sub-features are combined for next layer learning. 
Note that we utilize the simplest window method to split the features in this article. Please see the ablation results on the effect of the window size and number of sub-feature on the feature splitting scheme in Table \ref{tab:impact_of_window_size_and_number_of_sub-features}.

\subsubsection{Cell-embedded block}



Generally, traditional message passing (MP) mechanism can be regarded as a refinement on a discrete space, analogous to an interpolation operation, which implies that edges are essentially interpolated from nodes. A MP mechanism introducing the cell has potential to further enhance the refinement of the discrete space (namely, secondary refinement), thereby reducing the magnitude of discretization errors spatially and paving the way for its application in complex domains.

\begin{definition}[Cell in Graphs]
For a graph $G = (V, E)$, where $V$ is the set of nodes $\mathbf{v}$ and $E \subseteq V \times V$ is the set of edges, a cell in $G$ is a subset of nodes $C \subseteq V$, such that the nodes in $C$ form a complete subgraph (clique) or satisfy predefined structural relationships. In particular, a $k$-cell $C_k$ in a graph $G$ contains $k+1$ nodes. Consider $\forall (i,j,k,\dots) \in C_k$ and $ (\mathbf{e}_{ij},\mathbf{e}_{jk},\dots) \in E$, various structures could be formulated, such as node ($k=0$), edge ($k=1$), triangle ($k=2$), tetrahedron ($k=3$), and so on. Figure \ref{fig:cell} depicts the cell in graphs. 
\end{definition}

\begin{corollary}[Expressive Power]\label{corollary_1}
Given a graph $G$ including many $k$-cell ($k=0,1,2,\dots$), there exists a cell-based scheme that is more expressive than Weisfeiler-Lehman (WL) tests in distinguishing non-isomorphic graphs (see the proof in Supplementary \ref{wl_test}).
\end{corollary}

Therefore, we proposed a two-level cell-embedded mechanism to process the message on graphs. A detailed framework with cell-embedded features is designed as the following description.

\begin{remark}
Our method is geometry-agnostic, where the cell (e.g., triangle, quad, hexagon, tetrahedron) is not a limiting prior but rather input-adaptive guidance. The message-passing scheme learns geometric relations without shape-specific assumptions, ensuring universal applicability to diverse geometries as demonstrated in experiments spanning 2D/3D regular/irregular meshes.
\end{remark}

\subsubsection{Main Architecture}

\paragraph{\textbf{{Encoder}.}}
The encoder block maps the low-dimensional variables to corresponding high-dimensional latent features via differential functions (e.g., MLPs). The initial node feature $\mathbf{h}_{i}^{0}$ includes the node feature, one-hot feature of node type, and their position information. The initial edge feature $\mathbf{e}_{ij}^{0}$ contains the relative position vector, the distance of neighbor nodes, and etc. The initial cell feature $\mathbf{c}_{ijk}^{0}$ involves the centroid position of cell, the area of cell, and the relative position vector from three nodes to the corresponding centroid position. The corresponding forms are described as:
\begin{subequations}
\begin{align}
\mathbf{h}_{i}^{0}&=\phi_{v}^{en} \left(\mathbf{u}_{i} , \mathbf{x}_{i} , \kappa_{i} , \dots \right),\\
\mathbf{e}_{ij}^{0}&=\phi_{e}^{en} \left(\left(\mathbf{x}_{j}-\mathbf{x}_{i}\right) , d_{ij} , \dots \right),\\
\mathbf{c}_{ijk}^{0}&=\phi_{c}^{en} \left(\left(
\mathbf{x}_{\tau}-\mathbf{x}_{\triangle_{ijk}}\right)_{\tau \in \triangle_{ijk}},
\mathbf{x}_{\triangle_{ijk}}, A_{\triangle_{ijk}}, \dots \right), 
\end{align}
\end{subequations}
where the learnable functions $\phi_{v}^{en}(\cdot)$, $\phi_{e}^{en}(\cdot)$, and $\phi_{c}^{en}(\cdot)$ are applied to learn the latent features of node, edge, and cell; $(\mathbf{x}_{j}-\mathbf{x}_{i})$ a relative position vector between the nodes $i$ and $j$; $d_{ij}$ the relative physical distance; $\kappa_{i}$ the type of node $i$; the $\mathbf{x}_{\triangle_{ijk}}$ the centroid position of the cell $\triangle_{ijk}$; the $A_{\triangle_{ijk}}$ the area of the cell $\triangle ijk$. In addition, $(\cdot , \cdot)$ denotes the concatenation operation. 

\paragraph{\textbf{{Processor}.}}
The processor iteratively processes the latent features from the upstream encoder via the CE block. We divided the original edge channel $\mathbf{e}_{ij}^{l}$ into two parts: itself and its adjacent cells. With this simple process, the node can exchanges information with itself ({nodal info}), its immediate neighbor edges ({derivative info}), and its adjacent cells ({integral info}). In general, the cell features $\mathbf{c}^{l+1}_{ijk}$ and edge features $\mathbf{e}^{l+1}_{ij}$ firstly learn the effective information from adjacent nodes features, and then are aggregated to formulate the next node states $\mathbf{h}^{l+1}_{i}$. 

\textbf{CE block}: The two-level procedure is described as:
\begin{subequations}
\begin{align}
\mathbf{c}_{ijk}^{l+1}&=\phi_{c}^{l} \left (\mathbf{h}_{i}^{l},\mathbf{h}_{j}^{l},\mathbf{h}_{k}^{l},\mathbf{c}_{ijk}^{l} \right),\\
\mathbf{e}_{ij}^{l+1}&=\phi_{e}^{l} \left (\mathbf{h}_{i}^{l},\mathbf{h}_{j}^{l},\mathbf{e}_{ij}^{l} \right),\\
\mathbf{h}_{i}^{l+1}&=\phi_{v}^{l} \Big(\underbrace{\mathbf{h}_{i}^{l}}_{\text{nodal info}},\underbrace{\sum\nolimits_{j \in \mathcal{N}_{i}}\mathbf{e}_{ij}^{l+1}}_{\text{derivative info}},\underbrace{\sum\nolimits_{jk \in \mathcal{N}_{i}} \mathbf{c}_{ijk}^{l+1}}_{\text{integral info}} \Big),
\end{align}
\end{subequations}
where $j \in \mathcal{N}_{i}$ represents every neighbor edge $\mathbf{e}_{ij}$ at node $i$; $jk \in \mathcal{N}_{i}$ every neighbor cell $\triangle {ijk}$ at node $i$. $\phi_{c}^{l}, \phi_{e}^{l}$, and $ \phi_{v}^{l}$ are the differential functions of cell, edge, and node at the layer $l$. Note that we have reformulated the message passing mechanism, where edge and cell features, without interaction, are used to simultaneously update the node features. See the test results of computational cost and scalability in Table \ref{tab:computational_cost_and_scalability}.  

\textbf{FE block}: We apply the FE block on the above node features $\mathbf{h}_{i}^{l+1}$ to obtain the output node features of the processor that:
\begin{equation}
\mathbf{\bar{h}}_{i}^{l+1} = \phi^{fe}_{v}(\mathbf{h}_{i}^{l+1}),
\end{equation}
where $\phi_{v}^{fe}$ is the block mentioned in the Section \ref{subsubsec:fe}. A detailed diagram is shown in Figure \ref{fig:architecture}\textbf{c}.

\paragraph{\textbf{{Decoder}.}}
The decoder maps latent features back to physical variables on graphs. With a skip connection, we acquire new states $\mathbf{u}_{t_{k+1}}$ by incremental learning, described as: 
\begin{equation}\label{eq:decoder}
\hat{\mathbf{u}}_{i,t_{k+1}} =\phi_{v}^{de} \left(\mathbf{\bar{h}}_{i}^{L} \right)+\mathbf{u}_{i,t_{k}},
\end{equation}
where $\phi_{v}^{de}(\cdot)$ is the differentiable function (e.g., MLPs) and $L$ the total number of layers.

\vspace{-2pt}
\section{Experiments}\label{Experiments}

\subsection{Datasets and Baselines}
To evaluate the performance of CeFeGNN, we experiment on the classic physical problems and more challenging real-world scenarios, including Burgers equation, Gray-Scott Reaction-Diffusion (GS RD) equation, FitzHugh-Nagumo  (FN) equation, 
and Black-Sea (BS) dataset. The first three datasets are generated by various governing equations on the grid domain and the final one on irregular meshes. Here, the initial input fields of three synthetic datasets are generated on Gaussian distribution with various random seeds and the node connectivity is obtained by the Delaunay algorithm. See Appendix Table \ref{tab:Summary Information of Datasets} for a detailed description of these datasets. All PDE datasets in our paper are benchmarks and publicly available from \cite{brandstetter2022message,lienen2022learning,rao2023encoding} and the real-world dataset is collected by the CMEMS.  More details about dataset generation and source refer to Appendix Section \ref{Supplementary Details of Datasets}. Moreover, an extension of CeFeGNN on traffic prediction task is provided in Appendix section \ref{Extension on traffic prediction}.

We compare our model with the most popular graph-based neural network, such as MeshGraphNet (MGN) ~\cite{pfaff2021Learning}, Graph Attention Network  (GAT) ~\cite{velickovic2017graph}, Graph Attention Network Variant (GATv2) ~\cite{brody2021attentive}, 
the state-of-the-art models, message passing neural PDE solver  (MP-PDE) ~\cite{brandstetter2022message}  
, Fourier Neural Operator  (FNO) ~\cite{li2021Fourier}, Factorized FNO (FFNO) ~\cite{tran2021factorized}, Geometry-informed FNO (Geo-FNO) ~\cite{li2023fourier,li2024geometry}, Transolver~\cite{wu2024transolver}.
As shown in Appendix Table ~\ref{tab:Summary analysis of Baselines}, a comparative analysis of these baselines is discussed. Additional detailed information about these baselines refers to Appendix Section \ref{Supplementary Details of Baseline}.

\begin{figure*}[t!]
\centering
\includegraphics[width=0.96\textwidth]{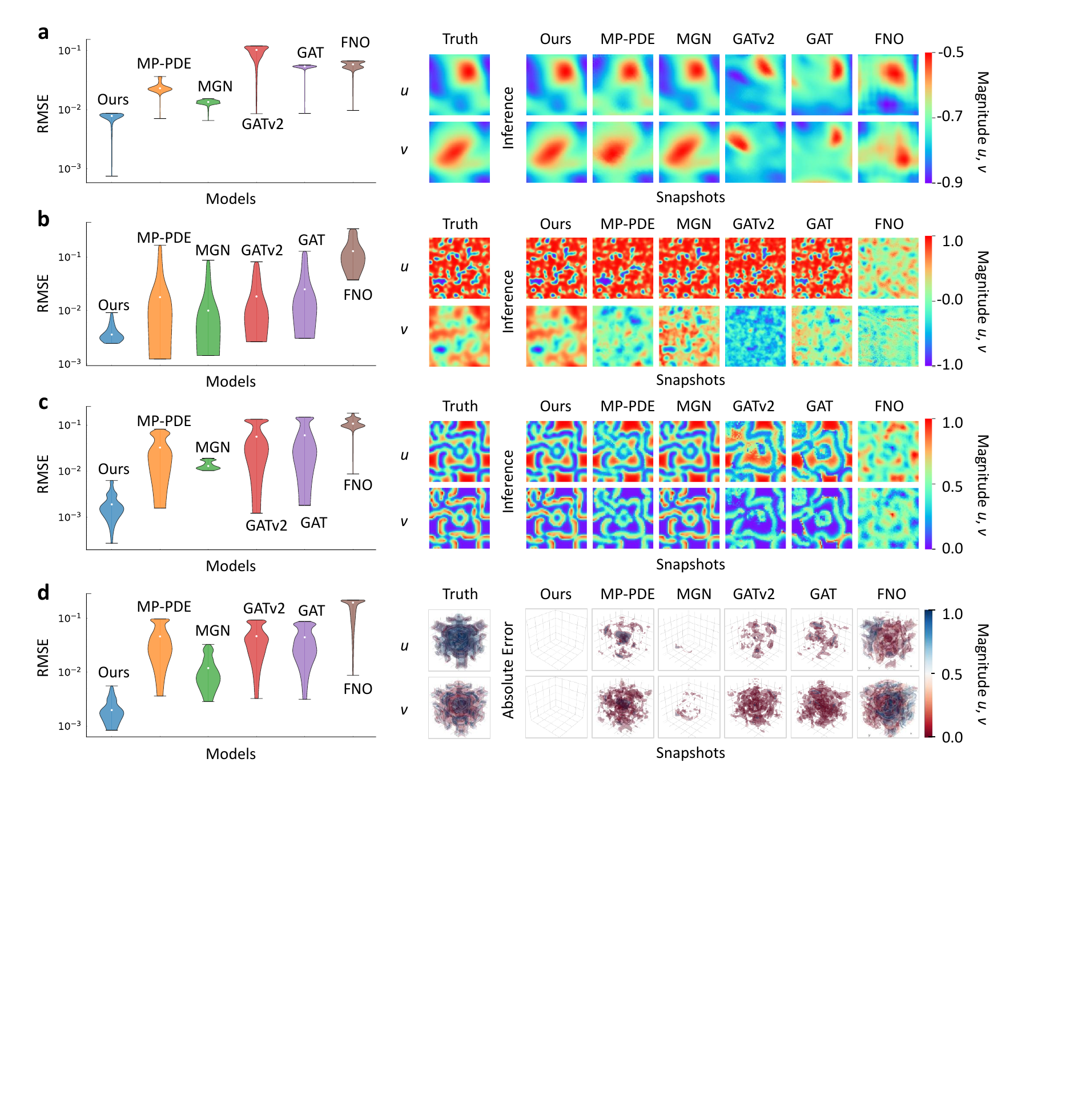}
\vspace{-6pt}
\caption{The test results of of all models on various datasets. \textbf{a-d}, the error distribution and the slices of generalization test on four grid-based datasets. The full snapshots of 3D GS dataset are displayed in Appendix Figure \ref{fig:snapshot_3dgs}.
The results of 2D BS dataset are displayed in Figure \ref{fig:bs_result}. 
Error propagation curves refer to Appendix Section \ref{Error propagation curve}.}
\label{fig:main_result}
\vspace{-6pt}
\end{figure*}

\subsection{Experimental setup}
In our experiments, we mainly focus on predicting much longer time steps with lower error and attempt to achieve better generalization ability of various initial conditions (ICs) and boundary conditions (BCs). For fairness, we set the feature dimension to 128 and utilize the one-step training strategy (i.e., one-step forward, one-step backward) for all tasks. All experiments are run on one NVIDIA A100 GPU. The Adaptive Moment Estimation (Adam) optimizer \cite{kingmaB2015adam} and the ReduceLROnPlateau learning scheduler \cite{Paszke2019PyTorch} with a decay rate of $0.8$ are utilized for the model training. All models are trained by injecting random Gaussian noise of varying standard deviation to the input to improve stability during rollout and correct small errors. Meanwhile, Root Mean Square Error (RMSE) loss function is utilized to optimize parameters $\theta$ in networks. Given the ground-truth values $Y \in \mathbb{R}^{N\times d}$ and the predictions $\hat{Y}\in \mathbb{R}^{N\times d}$ at time $t$, the loss function is defined as: 
$
    RMSE(Y, \hat{Y}) = \sqrt{\frac{1}{Nd} \sum\nolimits_{i=i}^{N} \sum\nolimits_{j=1}^{d}\left( y_{ij}-\hat{y}_{ij}\right)^{2}},
$
where $y_{ij}\in Y$ and $\hat{y}_{ij} \in \hat{Y}$. 
More details about the experimental setup are described in Appendix Section \ref{Supplementary Details of results}. 






\subsection{Results}

We consider four different types of study cases: (1) the generalization test, (2) the cell feature's efficacy, (3) the feature-enhanced effect, and (4) an ablation study. All our experiments revolve around the following question: \textit{\textbf{Can our model generalize well and achieve lower error with limited data?}}  

\subsubsection{Generalization test}
\begin{figure*}[t!]
\centering
\includegraphics[width=\textwidth]{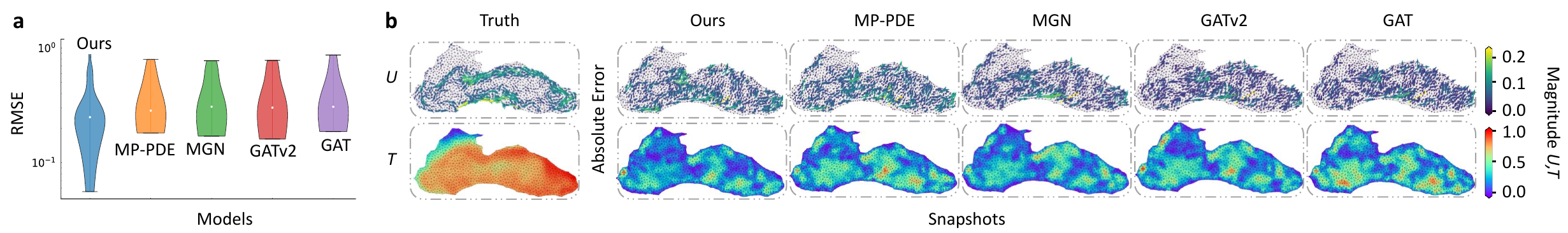}
\vspace{-16pt}
\caption{The test results of all models on BS Dataset. \textbf{a}, the error distribution. \textbf{b}, the prediction at the 10th time step. Error propagation curves refer to Appendix Section \ref{Error propagation curve}. The full snapshots on BS dataset refer to Appendix Figure \ref{fig:snapshot_2dbs}.}
\label{fig:bs_result}
\vspace{-6pt}
\end{figure*}

In this part, we varied the initial input field, randomly sampled from a Gaussian distribution with various means and standard deviations, in order to test the generalization ability of all models. According to the analysis results in Figures \ref{fig:main_result} and \ref{fig:bs_result}, we found that CeFeGNN generalizes to different ICs robustly on all benchmarks. It is evident that the performances of CeFeGNN and all baselines in the multi-step long-term prediction vary significantly. However, the experiment results of FNO show that it performs relatively poorly on all datasets except for the Burgers equation. 
This is a common observation \cite{mi2025conservation} for FNO on other benchmarks, especially with limited training data (e.g., a few to dozens of trajectories considered in this work, rather than thousands used in the original FNO paper). Under such a case, FNO struggles with over-fitting issue and fails to generalize over generic ICs for long-range prediction, especially in the 3D case. The full snapshots of 3D GS dataset are displayed in Appendix Figure \ref{fig:snapshot_3dgs}. The variant of FNO, FFNO, outperforms FNO across all datasets. Unexpectedly, Geo-FNO, which incorporates the IPHI technique, achieves the worst performance. Similarly, Transolver underperforms in all cases. Given the experimental results of above baselines, we can infer that these methods exhibit poor generalization ability on small datasets, falling short compared to graph-based methods.

\begin{table}[t!]
\caption{Results of different methods. ``--'' represents that the model is unable or unsuitable to learn the dynamics directly. ``$\downarrow$'' represents that the smaller the value of the quantitative metric, the better the model performance. The \textbf{bold} values and \underline{underlined} values represent the optimal and sub-optimal results on various datasets. The abbreviation ``Pro.'' represents promotion calculated from the above two. We describe the time interval with the form ``(1.0 s)''.
}
\label{tab:Performance Summary}
\vspace{-6pt}
\begin{center}
\begin{small}
\begin{tabular}{l|cccc|c}
\toprule
\multirow{4}{*}{\textbf{Model}}&\multicolumn{5}{c}{\textbf{RMSE $\downarrow$}}\\
\cmidrule{2-6}
& \makecell[c]{\textbf{2D Burgers} \\ (0.001 s)} 
& \makecell[c]{\textbf{2D FN}  \\ (0.002 s)} 
& \makecell[c]{\textbf{2D GS}  \\ (0.25 s)} 
& \makecell[c]{\textbf{3D GS}  \\ (0.25 s)}
& \makecell[c]{\textbf{2D BS}  \\ (1 day)}\\
\midrule
GAT &0.11754&0.02589&0.07227&0.06396&0.62954\\
GATv2 &0.11944&0.03827&0.07301&0.04519&0.64796\\
MGN &\underline{0.01174}&\underline{0.02108}&\underline{0.02917}&\underline{0.01925}&0.61475\\
MP-PDE&0.01784&0.02848&0.03860&0.06528&\underline{0.60761} \\
FNO &0.05754&0.12643&0.11331&0.17163& --\\
FFNO &0.03341&0.11921&0.03628&0.03594&--\\
Geo-FNO &0.59363&20.514&0.18669&NaN & 1.28931
\\
Transolver &0.17422&0.13724&0.18594&0.15204&0.81991\\
Ours &\textbf{0.00664}&\textbf{0.00364}&\textbf{0.00248}&\textbf{0.00138}&\textbf{0.55599}\\
\midrule
Pro. (\%) $\uparrow$ & 43.4&82.9&91.4&92.8&8.4\\
\bottomrule
\end{tabular}
\end{small}
\end{center}
\vspace{-12pt}
\end{table}

All results listed in Table \ref{tab:Performance Summary} demonstrate that all graph-based models have great generalization ability. GATv2, the advanced variant of GAT, underperforms GAT on Burgers, FN, 2D GS RD, and BS datasets, but outperforms GAT on 3D GS RD equation, yet both methods fall short of MGN. Surprisingly, the performance of MP-PDE is mediocre. Although MP-PDE is trained by the multi-step prediction strategy during the training stage, its results are only slightly better than MGN on the BS dataset. In contrast, our method performs robustly with much smaller errors in the multi-step prediction problem on all datasets. 

\subsubsection{The efficacy of cell features}
We also have test the efficacy of cell features on the performance of graph-based baselines across all benchmarks with various methods and report the results with RMSE metrics in the Table \ref{tab:Effect_cell} below to support our claim of introducing cells. We can intuitively see that the introduction of cell feature has a positive effect on all graph-based models, showing that our lightweight and decoupling alternative two-level cell-embedded message passing mechanism could better capture the spatial dependency on complex graphs. 
The consistent performance improvement (e.g., the promotions of ``MGN + Cell'' and ``MP-PDE + Cell'') not only exhibits strong practicability of cell features but also reflects a right breakthrough to improve the traditional ``black-box'' neural network via the prior knowledge (e.g., geometric concept) embedded mechanism.

\subsubsection{Feature-enhanced effect}
We investigate the effectiveness of the FE block on all graph-based network over all datasets. The results are reported in Table \ref{tab:Effect_fe}, showing that the feature-enhanced block somewhat changes the performance of these networks. We can directly see that this module has improved the performance of all baselines on real-world datasets. Specifically, it achieves the best and worst promotions in the cases of ``MGN + FE'' on the 3D GS RD equation and ``GATv2 + FE'' on the 2D Burgers equation. Intriguingly, after embedding the FE block in the processor block, the attention-based graph networks (e.g., GAT and GATv2) perform worse on the governing equation and better on the real-world dataset (even though the promotion is not large). 
This negative impact of the FE block on attention-based methods is essentially due to a logical conflict of design motivation. Consider two adjacent nodes $\mathbf{h}_1$ and $\mathbf{h}_2$ near a node $\mathbf{h}_0$, the attention mechanism assigns normalized weights $w_1$ and $w_2$ (e.g., $w_1+w_2 = 1$) and aggregates features by the summation operation like $w_1 \mathbf{h}_1+w_2 \mathbf{h}_2$. However, the FE block would disrupt this global normalization rule in the attention mechanism and reduces the expression capability of attention-based methods.


\begin{table}[t!]
\caption{Efficacy of cell features on the performance of graph-based baselines across all benchmarks.
}
\label{tab:Effect_cell}
\vspace{-6pt}
\begin{center}
\begin{small}
\begin{tabular}{l|cccc|c}
\toprule
\multirow{4}{*}{\textbf{Model}}&\multicolumn{5}{c}{\textbf{RMSE $\downarrow$}}\\
\cmidrule{2-6}
& \makecell[c]{\textbf{2D Burgers} \\ (0.001 s)} 
& \makecell[c]{\textbf{2D FN}  \\ (0.002 s)} 
& \makecell[c]{\textbf{2D GS }  \\ (0.25 s)} 
& \makecell[c]{\textbf{3D GS }  \\ (0.25 s)}
& \makecell[c]{\textbf{2D BS}  \\ (1 day)}\\
\midrule
MGN &0.01174&0.02108&0.02917&0.01925&0.61475\\
 + Cell &0.00826&0.00791&0.00832&0.00694&0.58019\\
\midrule
Pro. (\%) $\uparrow$ &29.6&62.4&71.4&63.9&5.6\\
\midrule
MP-PDE&0.01784&0.02848&0.03860&0.06528&0.60761\\
 + Cell &0.00951&0.01193&0.00947&0.00992&0.59313\\
\midrule
Pro. (\%) $\uparrow$ &46.7&58.1&75.4&84.8&2.38\\
\bottomrule
\end{tabular}
\end{small}
\end{center}
\vspace{-12pt}
\end{table}

\begin{table}[t!]
\caption{Efficacy of the FE block on the performance of graph-based baselines across all benchmarks. 
}
\label{tab:Effect_fe}
\vspace{-6pt}
\begin{center}
\begin{small}
\begin{tabular}{l|cccc|c}
\toprule
\multirow{4}{*}{\textbf{Model}}&\multicolumn{5}{c}{\textbf{RMSE $\downarrow$}}\\
\cmidrule{2-6}
& \makecell[c]{\textbf{2D Burgers} \\ (0.001 s)} 
& \makecell[c]{\textbf{2D FN}  \\ (0.002 s)} 
& \makecell[c]{\textbf{2D GS}  \\ (0.25 s)} 
& \makecell[c]{\textbf{3D GS}  \\ (0.25 s)}
& \makecell[c]{\textbf{2D BS}  \\ (1 day)}\\
\midrule
GAT &0.11754&0.02589&0.07227&0.06396&0.62954\\
 + FE &0.15132&0.02717&0.08527&0.07058&0.61984\\
\midrule
Pro. (\%) $\uparrow$ &$-$28.7&$-$4.9&$-$17.9&$-$10.3&1.5\\
\midrule
GATv2 &0.11944&0.03827&0.07301&0.04519&0.64796\\
 + FE &0.18496&0.04117&0.09365&0.06432 &0.63363\\
\midrule
Pro. (\%) $\uparrow$ &$-$54.8&$-$7.5&$-$28.2&$-$43.2&2.2\\
\midrule
MGN &0.01174&0.02108&0.02917&0.01925&0.61475\\
 + FE &0.00817&0.01241&0.01583&0.00721&0.60593\\
\midrule
Pro. (\%) $\uparrow$ &30.4&41.1&45.7&62.5&1.4\\
\midrule
MP-PDE&0.01784&0.02848&0.03860&0.06528&0.60761\\
 + FE &0.01445&0.01957&0.02621&0.03655&0.60372\\
\midrule
Pro. (\%) $\uparrow$ &18.9&31.2&32.1&41.5&0.6\\
\bottomrule
\end{tabular}
\end{small}
\end{center}
\vspace{-16pt}
\end{table}

\subsubsection{Ablation study}
In this part, we perform an ablation study on all datasets to assess the contributions of the FE and CE blocks in CeFeGNN, as shown in Table \ref{tab:Ablation Study}. The results indicate that, without the introduction of the cell, the assembly of the traditional node-edge message passing mechanism and FE block still shows good generalization ability, even with small data. Although we attempt to make the model learn higher-order information in each round, the results demonstrate that the cell-embedded mechanism did not achieve the desired performance, which means that the network architecture of traditional MPNNs still has the over-smoothness problems that need a better solution.

\begin{table}[t!]
\caption{Quantitative results of ablation study on CeFeGNN. The abbreviation ``C'' represents the cell.}
\label{tab:Ablation Study}
\vspace{-6pt}
\begin{center}
\begin{small}
\begin{tabular}{l|cccc|c}
\toprule
\multirow{4}{*}{\textbf{Model}}&\multicolumn{5}{c}{\textbf{RMSE $\downarrow$}}\\
\cmidrule{2-6}
& \makecell[c]{\textbf{2D Burgers} \\ (0.001 s)} 
& \makecell[c]{\textbf{2D FN}  \\ (0.002 s)} 
& \makecell[c]{\textbf{2D GS}  \\ (0.25 s)} 
& \makecell[c]{\textbf{3D GS}  \\ (0.25 s)}
& \makecell[c]{\textbf{2D BS}  \\ (1 day)}\\
\midrule
\textbf{w/o} C, FE   & 0.01174 & 0.02108 & 0.02917 & 0.01925 & 0.61475 \\
\textbf{w/o} C  &\underline{0.00828}&0.00982&{0.01035}&{0.00680}&\underline{0.58236}     \\
\textbf{w/o} FE    &0.00877&\underline{0.00788}&\underline{0.00803}&\underline{0.00679} &0.58271     \\
\midrule
Ours    &\textbf{0.00664}&\textbf{0.00364}&\textbf{0.00248}&\textbf{0.00138}&\textbf{0.55599}     \\
\bottomrule
\end{tabular}
\end{small}
\end{center}
 \vspace{-6pt}
\end{table}

\begin{table}[t!]
\caption{Quantitative results of computational cost and scalability of CeFeGNN. Note that the dimension of latent feature is set as 128. The abbreviation ``Param.'' represents the number of parameter. ``L'' is the total number of processor.}
\label{tab:computational_cost_and_scalability}
\vspace{-6pt}
\begin{center}
\begin{small}
\begin{tabular}{lccccc}
\toprule
\multirow{2}{*}{\textbf{Model}}
&\multirow{2}{*}{\makecell[c]{\textbf{L}}}
&\multirow{2}{*}{\makecell[c]{\textbf{Param.}}}
&\multirow{2}{*}{\makecell[c]{\textbf{Training}\\ {(s/epoch)}}}
&\multirow{2}{*}{\makecell[c]{\textbf{GPU usage} \\(GB)}}
&\multirow{2}{*}{\makecell[c]{\textbf{2D Burgers} $\downarrow$ \\ (0.001 s)} }\\
\\
\midrule
MGN          &4      &514,236   &8.64  & 32.56  &0.01174\\ 
MGN          &12     &1,438,086 &20.90 & 74.81  &0.01858\\
\midrule
MP-PDE       &4      &528,658   &7.86  & 33.23  &0.01784\\ 
MP-PDE       &12     &1,230,533 &17.44 & 76.63  &0.10101\\
\midrule
\textbf{w/o} FE       &4      &971,916   &14.86 & 45.85  &\underline{0.00877}\\
Ours        &4      &1,482,674 &17.31 & 45.87  &\textbf{0.00664}\\
\bottomrule
\end{tabular}
\end{small}
\end{center}
 \vspace{-6pt}
\end{table}

\begin{table}[t!]
\caption{Quantitative results of the impact of window size and number of sub-features of CeFeGNN. Note that the number for window size and sub-feature is described by the form like ``(1/128)''. The abbreviation ``Param.'' represents the number of parameter.}
\label{tab:impact_of_window_size_and_number_of_sub-features}
\vspace{-4pt}
\begin{center}
\begin{small}
\begin{tabular}{cc|cccc}
\toprule
\multirow{4}{*}{\makecell[c]{\textbf{Case}}}
&\multirow{4}{*}{\makecell[c]{\textbf{Param.}}}
&\multicolumn{4}{c}{\textbf{RMSE $\downarrow$}}\\
\cmidrule{3-6}
&& \makecell[c]{\textbf{2D Burgers} \\ ( 0.001 s)} 
& \makecell[c]{\textbf{2D FN}  \\ (0.002 s)} 
& \makecell[c]{\textbf{2D GS }  \\ (0.25 s)} 
& \makecell[c]{\textbf{3D GS }  \\ (0.25 s)}\\
\midrule
(1/128) &9,362,572&\underline{0.00665}&0.00409&0.00251&\textbf{0.00122}\\
(2/64)  &3,071,116&0.00714&\underline{0.00370}&\textbf{0.00232}&0.00192\\
(4/32)&1,482,674&\textbf{0.00664}&\textbf{0.00364}&\underline{0.00248}&\underline{0.00138}\\
(8/16)  &1,105,036&0.01263&0.02261&0.03481&0.04128\\
(16/8)   &1,006,736&0.11753&0.12900&0.09120&0.18963\\
\bottomrule
\end{tabular}
\end{small}
\end{center}
\vspace{-6pt}
\end{table}

As shown in Table \ref{tab:computational_cost_and_scalability}, our model with similar parameter ranges achieves the best performance compared with MGN and MP-PDE. We have verified that the improvement of our model performance is due to the innovative use of cell features rather than the introduction of additional parameters.
For precision-oriented tasks (e.g., modeling PDEs), some trade-off in terms of speed while achieving much higher precision is acceptable. For example, compared to suboptimal models, CeFeGNN achieves a RMSE reduction of 10 $\times$ or more (see Table \ref{tab:Performance Summary}) on FN and GS datasets. Even without the FE module (see Table \ref{tab:Ablation Study}), our model still yields a 3
$\times$ RMSE reduction.

In addition, as shown in Table \ref{tab:impact_of_window_size_and_number_of_sub-features}, excessive feature segmentation can disrupt feature correlations and degrade model performance. Thus, we use the splitting method with a proper window size. We also have performed a data scaling test and report the results (data size vs. prediction error) in the Appendix Table \ref{tab:data_scaling_results}.
The tests were conducted on the Burgers example using varying trajectories as training data. It can be observed that our model with a smaller amount of data has equal or superior performance compared with that of other methods (MGN and MP-PDE) with larger amounts of data. Additionally, we have investigated the effectiveness of the cell feature on graph-based networks over all benchmarks. The results in Table \ref{tab:Effect_cell} show the positive efficacy of cell features.
A ablation test about the relative cell position information also demonstrates cell's significance, shown in Appendix Table \ref{tab:cell_pos}.
More importantly, the comparison results between two-level and three-level message passing mechanisms in Appendix Table \ref{tab:message_level} have verified the he rationality of model design motivation.

\vspace{-2pt}
\section{Conclusion}\label{Conclusion}
In this paper, we proposed an end-to-end graph-based framework (namely, CeFeGNN) to learn the complex spatiotemporal dynamics. By utilizing the regional information, CeFeGNN predicts future long-term unobserved states and addresses the over-smoothness problem in GNNs. Firstly, the learnable cell attribution in CE block captures the spatial dependency of regional features, upgrading the local aggregation scheme from the first order to a higher order. Secondly, the FE block enriches the node features, maintaining strong representational power even after multiple rounds of aggregation. 
The superior performance of our model has been demonstrated by extensive experiments. The theoretical analysis (see Section \ref{Methodology}, Appendix Section \ref{Proofs}) and experimental validation (see Section \ref{Experiments}, Appendix Section \ref{Supplementary Details of Experiments}) differentiate our approach from existing models.
Although CeFeGNN achieves superior performance on extensive experiments, there are several directions for future work, including that (1) pushing our model to learn on a finer mesh with more complex boundary conditions, and (2) further exploring the potential of geometric information to learn higher-order attributions in a more refined way, rather than the rough handling in our article. We attempt to accomplish these goals in our future research work.



\bibliographystyle{ACM-Reference-Format}
\bibliography{reference}


\renewcommand{\thefigure}{S\arabic{figure}}
\setcounter{figure}{0} 

\renewcommand{\theequation}{S\arabic{equation}}
\setcounter{equation}{0} 

\renewcommand{\thetable}{S\arabic{table}}
\setcounter{table}{0} 

\appendix

\section{Variables notation}
In this part, we present a summary of the variable notations used in our paper, as detailed in Table \ref{tab:notations}.

\begin{table*}[t!]
\caption{Summary of Notations.}
\label{tab:notations}
\vspace{-4pt}
\begin{center}
\begin{tabular}{lll}
\toprule
\textbf{Calculus} &\textbf{Short Name}&\textbf{Role} \\
\midrule
Derivative of $y$ with respect to $x$&${d y}/{d x}$&Unknown Parameter\\
Partial derivative of $y$ with respect to $x$&${\partial y}/ {\partial x}$&Unknown Parameter\\
Gradient of $y$ with respect to $x$&$\nabla_x y $&Unknown Parameter\\
Tensor containing derivatives of $y$ with respect to $\mathbf{X}$ &$\nabla_\mathbf{X} y$&Unknown Parameter\\
Jacobian matrix $\mathbf{J} \in \mathbb{R}^{m\times n}$ of $f: \mathbb{R}^n \rightarrow \mathbb{R}^m$&${\partial f}/{\partial x}$&Unknown Parameter\\
The Hessian matrix of $f$ at input point $x$&$\nabla_x^2 f(x)\text{ or }\mathbf{H}(f)(x)$&Unknown Parameter\\
Definite integral over the entire domain of $x$&$\int f(x) dx$&Unknown Parameter\\
Definite integral with respect to $x$ over the set $S$ &$\int_S f(x) dx$&Unknown Parameter\\
\midrule
\textbf{Sets and Graphs} &\textbf{Short Name}&\textbf{Role} \\
\midrule
The set containing 0 and 1&$\{0, 1\}$&Predefined Variable\\
The set of all integers between $0$ and $n$&$\{0, 1, \dots, n \}$&Predefined Variable\\
The real interval including $a$ and $b$&$[a, b]$&Predefined Variable\\
The real interval excluding $a$ but including $b$&(a, b]&Predefined Variable\\
A graph with nodes and edges& $G=(E,V)$&Predefined Variable\\
\midrule
\textbf{Latent Variables} &\textbf{Short Name}&\textbf{Role} \\
\midrule
The node index & $i,j,k$&Predefined Variable \\
The latent node features &$\mathbf{h}_{i}$&Predefined Variable \\
The latent edge features &$\mathbf{e}_{ij}$&Predefined Variable \\
The latent cell features &$\mathbf{c}_{ijk}$&Predefined Variable \\
\midrule
\textbf{Variables} &\textbf{Short Name}&\textbf{Role} \\
\midrule
x-component of velocity & $u(\mathbf{x},t)$&Input/Predicted Variable \\
y-component of velocity & $v(\mathbf{x},t)$&Input/Predicted Variable \\
vorticity & $w(\mathbf{x},t)$&Input/Predicted Variable \\
pressure & $p(\mathbf{x},t)$&Input/Predicted Variable \\
temperature under water & $T(\mathbf{x},t)$&Input/Predicted Variable \\
\midrule
\textbf{Space and Time} &\textbf{Short Name}&\textbf{Role} \\
\midrule
x-direction of space coordinate & $x$ &Predefined Variable\\
y-direction of space coordinate & $y$&Predefined Variable \\
z-direction of space coordinate & $z$ &Predefined Variable\\
time coordinate & $t$ &Predefined Variable\\
time increment  & $\Delta t$ &Predefined Variable\\
space increment & $\Delta \mathbf{x}$&Predefined Variable \\
discrete timestamp at $k$th step & $t_{k}$&Predefined Variable \\
\bottomrule
\end{tabular}
\end{center}
\end{table*}

\section{Background: OOD test in Spatiotemporal Prediction}

In this part, we would like to clarify the generalization tests with different random ICs in spatiotemporal dynamics are OOD. For a nonlinear PDE system, even if the ICs are IID, the spatiotemporal solution can be OOD, because of the following reasons.

\paragraph{\textbf{{Nonlinearity and Emergence of Complex Patterns}.}}
Nonlinear PDEs are characterized by their ability to produce highly complex behavior over time. This nonlinearity can amplify small differences in ICs, leading to the emergence of patterns or behaviors that are vastly different from what was initially expected. Even if the ICs are IID, the interactions dictated by the nonlinear terms in the PDE can cause the solution to evolve in a way that is not reflective of the initial distribution. As a result, the system may exhibit behaviors that are not represented in the original distribution, leading to an OOD trajectory dataset. For example, in the Burgers and FN examples, the ICs are generated based on Gaussian distribution with different random seeds (e.g., IID); however, the corresponding solution trajectories remain OOD judging from the histogram plots.

\paragraph{\textbf{{Chaotic Dynamics}.} }
Nonlinear PDEs may exhibit chaotic behavior. In these systems, small perturbations in ICs can lead to exponentially different solutions. Over time, this chaotic evolution can cause the solution to become highly sensitive to ICs (e.g., the 2D FN and 2D/3D GS RD test data examples shown in Figure \ref{fig:examples}b-d), resulting in a distribution that is very different from the IID distribution of ICs, yielding an OOD solution space.

\paragraph{\textbf{{Long-term Evolution}.}}
In many nonlinear PDEs, solutions tend to evolve toward certain stable structures or steady states known as attractors. These attractors can be complex structures in the solution space. Over time, the solution might converge to or oscillate around these attractors, regardless of the IID nature of ICs (e.g., the Burgers, FN and GS RD examples in our paper). The distribution of solutions near these attractors can be very different from the IC IID distribution. Essentially, the system's long-term behavior is determined more by the attractors rather than by ICs.

\paragraph{\textbf{{Spatialtemporal Correlations}.}}
The assumption of IID ICs implies no spatial/temporal correlations initially. However, the evolving dynamics governed by PDEs can introduce correlations over time. These correlations can lead to a solution that has a distribution quite different from the original IID distribution. The emergence of such correlations indicates that the evolved solution is not just a simple extension of ICs, producing datasets with OOD.

\paragraph{\textbf{{Breaking of Statistical Assumptions}.}}
As the system evolves, the assumptions that justified the IID nature of ICs may no longer hold. The dynamics of the PDE can induce structures, patterns, or dependencies that were not present in ICs. As a result, the statistical properties of the evolved solution may diverge from those of ICs, leading to an OOD.

Hence, even though the ICs are IID, the resulting solution trajectories are OOD. This might be a little bit different from our understanding of IID/OOD datasets in common practices of NLP, CV, etc. In addition, we have indeed considered OOD ICs in our tests, e.g., the ICs of the 2D/3D GS RD  examples are randomly placed square cube concentrations (1 or 2 square/cube blocks) as shown in Figure \ref{fig:examples}c-d. The resulting solution datasets are obviously OOD. Therefore, all the results of generalizing to different ICs represent OOD tests. Given the same comparison test sets, our model shows better generalization performance over other baselines.

\section{Proofs}\label{Proofs}

\subsection{Supplementary Definitions, Lemmas, Theorems, Corollaries, or Proofs}\label{proof_1}

\begin{lemma}[Feature Diversity]
Introducing cells enhances feature diversity by encoding higher-order relationships among nodes. Specifically, the basic features in traditional MP mechanism are $\{ \mathbf{h}_i$,$ \mathbf{h}_j$,$\mathbf{e}_{ij}$, \\ $ \mathbf{e}_{ji} \}$, and the basic features in cell-based MP mechanism are $\{ \mathbf{h}_i$, $\mathbf{h}_j, \mathbf{h}_k, \mathbf{e}_{ij}$, $ \mathbf{e}_{jk}, \mathbf{e}_{ki}, \mathbf{c}_{ijk}, \mathbf{c}_{kij}, \mathbf{c}_{jki}\}$. The additional node $\mathbf{h}_k$ provides a richer context, enabling the capture of more complex patterns within one round.
\end{lemma}

\begin{lemma}[Reduction in Ambiguity]
Traditional MP methods rely solely on pairwise interactions, which can lead to ambiguity in cases of structural symmetry. Cell-based MP mechanism leverages the higher-order structure, reducing ambiguity by providing additional constraints through relationships between three or more nodes.
\end{lemma}

\begin{corollary}[Improved Performance]
Cell-based MP mechanism improves prediction capability by enhancing feature distinguishability through introducing higher-order relationships.
\end{corollary}

\begin{corollary}[Suitability for Graphs] 
In dense or sparse graphs, cell-based methods outperform traditional methods by capturing multi-node interactions within one round, which are critical for preserving the graph's topology.
\end{corollary}

\begin{remark}[Explanation of Corollary \ref{corollary_1} in our main paper]
1-WL test relies on pairwise node comparisons and cannot distinguish graphs that are symmetric under pairwise relationships. By lifting graphs to a cell-embedded pattern and using cell-based MP mechanism, higher-order interactions are encoded, allowing discrimination of graphs that 1-WL test cannot separate. See the detail proof in Subsection \ref{wl_test}.
\end{remark}

\subsection{Weisfeiler-Lehman (WL) Tests}\label{wl_test}
Weisfeiler-Lehman (WL) Test is an iterative graph isomorphism algorithm that updates node features by aggregating the features of neighboring nodes. After each iteration, the updated node features are hashed to encode structural information. Despite its effectiveness, WL test cannot distinguish certain non-isomorphic graphs, particularly when higher-order structural information is required. In this part, our goal is to show that cell-based message passing mechanism is more expressive than the 1-WL test for distinguishing non-isomorphic graphs.

\paragraph{\textbf{Task Definition.}}
A graph $G = (V, E)$ has a set of vertices $V$ and edges $E$. The 1-WL test iteratively computes node features $\mathbf{h}_i^{l}$ at iteration $l$ as $ \mathbf{h}_i^{l+1} = \text{Hash}\left(\mathbf{h}_i^{l}, \{\mathbf{h}_j^{l} : j \in \mathcal{N}_i\}\right)$, where $\mathcal{N}_i$ is the set of neighbors of $i$, and $\mathbf{h}_i^{0}$ is initialized with the node's feature.

\begin{figure}[t!]
\centering
\includegraphics[width=0.6\columnwidth]{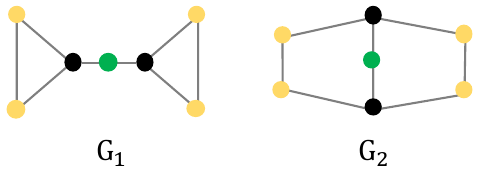}
\vspace{-6pt}
\caption{Two undistinguished graphs by 1-WL test. Different colors represent different labels.
}
\label{fig:non_isomorphic_graph}
\end{figure}

Consider two graphs $G_1$ and  $G_2$ (see Figure \ref{fig:non_isomorphic_graph}), the WL test fails to distinguish $G_1$ and $G_2$ because it only aggregates local neighborhood information, and both graphs have identical degree distributions and neighborhood structures for all nodes.

Given that cells (e.g., triangles) can be explicitly considered, we propose a simple cell-based scheme, described as follows.

\paragraph{\textbf{Initialization.}}
Initializing node features $\mathbf{h}_i^{0}$ based on their labels or degrees and initializing higher-order cell features $\mathbf{c}_{ijk}^{0}$ by aggregating node features within the triangle.

\paragraph{\textbf{Message Passing.}} Updating node and cell features iteratively. For example:
   \begin{subequations}
   \begin{align}
       \mathbf{h}_i^{l+1} &= \text{Hash}\left(\mathbf{h}_i^{l}, \{\mathbf{h}_j^{l} : j \in \mathcal{N}_i\}, \{\mathbf{c}_{ijk}^{l} : jk \in \mathcal{N}_i\}\right),\\
       \mathbf{c}_{ijk}^{l} &= \text{Aggregate}\left(\mathbf{h}_i^{l}, \mathbf{h}_j^{l}, \mathbf{h}_k^{l}\right), \quad \forall i,j,k \in \Delta_{ijk}.
   \end{align}
   \end{subequations}

\paragraph{\textbf{Expressiveness.}}
In $G_2$, no triangles exist, so all triangle-related features $\mathbf{c}_{ijk}$ will remain zero or absent. (2) In $G_1$, there are two triangles. These triangles generate non-zero features that propagate back to nodes during message passing. Thus, the presence of higher-order structures (triangles) allows the cell-based scheme to distinguish $G_1$ from $G_2$. And we proposed the following proposition about the expressiveness of cell-based message-passing scheme.

\begin{proposition}
The cell-based message-passing scheme is more expressive than 1-WL test because it captures higher-order interactions (e.g., triangles) that are invisible to 1-WL. This enhanced capability enables it to distinguish graphs, such as $G_1$ and $G_2$, which cannot be differentiated by 1-WL test.
\end{proposition}


\subsection{Inspiration of Feature-enhanced (FE) Block}\label{Mathematical Inspiration of Feature-enhanced (FE) Block}

The process of FE block is inspired by interaction models in physics and mathematically hypothesized to capture nonlinear dependencies, enhancing the model's representation power.

\paragraph{\textbf{Hypothesis.}}
The FE block's hypothesis could be framed as: By capturing second-order interactions between latent features and applying selective filtering, the model can better represent complex structures or relationships in the data.

\paragraph{\textbf{Physical Analogy.} }
In physics, the outer product and second-order terms are often used to model interactions, such as stress tensors in mechanics or pairwise correlations in quantum mechanics. Here, the module could draw an analogy to systems where interactions between individual components (features) are crucial to the overall behavior.

\paragraph{\textbf{Process Overview.}}
In detail, it regards the node latent feature ${\mathbf{\bar{h}}}_{i} \in \mathbb{R}^{D\times 1}$ as basis and builds a higher-order tensor feature $\mathbf{H}_{i} \in \mathbb{R}^{D\times D}$ via an outer-product operation, e.g., ${\mathbf{\bar{h}}}_{i}\otimes{\mathbf{\bar{h}}}_{i}$. This process in Algorithm \ref{alg:fe_algorithm} creates abundant second-order nonlinear terms to enrich the feature map. We then use a mask operation with $\mathbf{M} \in \mathbb{R}^{D\times D}$ to randomly sample these terms, filtering the appropriate information by a learnable weight tensor $\mathbf{W} \in \mathbb{R}^{D\times D\times D}$ to enhance the model's representation capacity.

\begin{algorithm}[htbp]
\caption{Feature-enhanced block}
\label{alg:fe_algorithm}
\KwIn{The current node states $\mathbf{h}^{l} \in \mathbb{R}^{N\times D}$ }
\textbf{Parameter:} The weight tensor $\mathbf{W}^{l} \in \mathbb{R}^{D\times D\times D}$ and the mask matrix $\mathbf{M}^{l} \in \mathbb{R}^{D\times D}$\\
\KwOut{The updated node states $ \hat{\mathbf{h}}^{l} \in \mathbb{R}^{N\times D}$}
\While{stop condition is not reached}{
 \textbf{Step 1: Reshape phase}. Reshape the input states $\mathbf{h}^{l}$ and obtain the ${\mathbf{\bar{h}}}^{l} \in \mathbb{R}^{N\times D\times 1}$.\\
 \textbf{Step 2: Expansion phase}. Play an Outer product operation on the ${\mathbf{\bar{h}}}^{l}$ to get new states $\mathbf{H}^{l} \in \mathbb{R}^{N\times D\times D}$.\\
 \textbf{Step 3: Filtering phase}. Filter feature with a Hadamard product operation on the new states $\mathbf{H}^{l}$ and the mask $\mathbf{M}^{l}$ to get new states $\hat{\mathbf{H}}^{l} \in \mathbb{R}^{N\times D\times D}$.\\
 \textbf{Step 4: Contraction phase}. Contract the new states $\hat{\mathbf{H}}^{l}$ with the weight $\mathbf{W}^{l}$ by a Double contraction operation to construct the final states $\tilde{\mathbf{h}}^{l} \in \mathbb{R}^{N\times D}$.\\
}
\textbf{return} The updated states $\tilde{\mathbf{h}}^{l}$.
\end{algorithm}




\section{Supplementary details of experiments}
\label{Supplementary Details of Experiments}

\subsection{Supplementary details of datasets}
\label{Supplementary Details of Datasets}

\begin{table*}[t!]
\caption{{Summary Information of Datasets.} Note that the trajectory number for training, validation, and testing is described by the form like (5/2/3).}
\label{tab:Summary Information of Datasets}
\vspace{-4pt}
\begin{center}
\begin{tabular}{lcccccccc}
\toprule
\makecell[l]{\textbf{Dataset}}
& \makecell[c]{\textbf{Space}\\\textbf{Interval}}  
& \makecell[c]{\textbf{Time}\\{ \textbf{Interval}}} 
& \makecell[c]{\textbf{PDE} \\ \textbf{Formulation}}  
& \makecell[c]{\textbf{Boundary} \\ \textbf{Condition}} 
& \makecell[c]{\textbf{Number of} \\ \textbf{Trajectories}}
& \makecell[c]{\textbf{Number of} \\ \textbf{Nodes} }
&\makecell[c]{\textbf{Trajectory} \\ \textbf{Length}}
& \makecell[c]{\textbf{Force} \\ \textbf{Term}} \\
\midrule
\makecell[l]{2D Burgers }&$\Delta x = $ 0.02&$\Delta t = $ 0.001 s & Eq. \ref{eq:burgers} &Periodic &  40 (30/5/5) &2,500 ($50^{2}$)&1000& No\\
\midrule
\makecell[l]{2D FN }&$\Delta x = $ 1&$\Delta t = $ 0.002 s & Eq. \ref{eq:fn} &Periodic &  10 (5/2/3) &16,384 ($128^{2}$)&3000& No \\
\midrule 
\makecell[l]{2D GS RD }&$\Delta x = $ 2&$\Delta t = $ 0.25 s & Eq. \ref{eq:gs} &Periodic & 10 (5/2/3) & 2304  ($48^{2}$)& 3000 & No\\
\midrule
\makecell[l]{3D GS RD }&$\Delta x = $ 2&$\Delta t = $ 0.25 s & Eq. \ref{eq:gs} &Periodic & 6 (2/2/2) & 13,824  ($24^{3}$)& 3000 & No\\
\midrule
\makecell[l]{2D BS }&N/A&$\Delta t = $ 1 day & -- & N/A&  24 (20/2/2) &1,000-20,000&365& N/A\\
\bottomrule
\end{tabular}
\end{center}
\vspace{-4pt}
\end{table*}

\paragraph{\textbf{{Viscous Burgers Equation}.}}
As a simple non-linear convection–diffusion PDE, Burgers equation generates fluid dynamics on various input parameters. For concreteness, within a given 2D field, there is a velocity $\mathbf{u}=[u, v]^{T}$ at per 2D grid point. Its general formulation is expressed as following description:
\begin{equation}
\label{eq:burgers}
\begin{aligned}
{\partial \mathbf{u}}/{\partial t} + \mathbf{u} \cdot \nabla \mathbf{u} = \mathbf{D} \nabla^{2}{\mathbf{u}},
\end{aligned}
\end{equation}
where viscosity $\mathbf{D} = [D_u, D_v]$ is the diffusion coefficient of fluid, $\mathbf{u} (\mathbf{x},t)$ its velocity, $\mathbf{x}$ the 2D spatial coordinate and $t$ the 1D temporal coordinate.
In this work, we generate the simulation trajectory within $\Omega \in [0,1]^{2}$ and $t \in[0, 1] (s)$, using a $4$th-order Runge–Kutta time integration method~~\cite{ren2022phycrnet} under the periodic condition.
Here, we define $D_u = 0.01, D_v = 0.01, \Delta t = 0.001 (s)$ and $\Delta x = 0.02$.

\paragraph{\textbf{{Fitzhugh-Nagumo Equation}.}}
Fitzhugh-Nagumo equation consists of a non-linear diffusion equation with different boundary conditions. We generates the training and testing data $\mathbf{u}=[u, v]^{T}$ in a given 2D field with periodic boundary conditions:
\begin{subequations}
\label{eq:fn}
\begin{align}
{\partial u (\mathbf{x},t)}/{\partial t} &= D_u \nabla^2 u + u - u^{3}-v^{3}+ \alpha,\\
{\partial v (\mathbf{x},t)}/{\partial t} &= D_v \nabla^2 v + (u-v) \times \beta,
\end{align}
\end{subequations}
where $D_u, D_v$ are the diffusion coefficients, $\alpha, \beta$  the reaction coefficients. 
In this work, we generate the simulation trajectory within $\Omega \in [0,128]^{2}$ and $t \in[0, 6] (s)$, using a $4$th-order Runge–Kutta time integration method~~\cite{ren2022phycrnet} under the periodic condition.
Here, we define $D_u = 1, D_v = 100, \alpha = 0.01, \beta = 0.25, \Delta t = 0.002 (s)$ and $ \Delta x = 1$. 



\paragraph{\textbf{{Gray-Scott Equation}.}}
As a coupled reaction-diffusion PDE, Gray-Scott equation consists of a velocity $\mathbf{u}=[ u, v]^{T}$. For example, given a 3D field, the corresponding form of each component on $\mathbf{x}= (x,y,z) \in\mathbb{R}^{3}$ and $t \in[0, T]$ is as follows:
\begin{subequations}
\label{eq:gs}
\begin{align}
{\partial {u (\mathbf{x},t)}}/{\partial t} &= D_u\nabla^2 {u} - {u}  {v}^{2} + \alpha   (1- {u}),\\
{\partial {v (\mathbf{x},t)}}/{\partial t} &= D_v\nabla^2{v} +{u} {v}^{2}- (\beta+\alpha) {v},
\end{align}
\end{subequations}
where $D_u$ and $D_v$ are the variable diffusion coefficients,
$\beta$ the conversion rate, $\alpha$ the in-flow rate of $u (\mathbf{x},t)$ from the outside, and $ (\alpha + \beta)$ the removal rate of $v (\mathbf{x},t)$ from the reaction field.
In this work, we generate the 2D simulation trajectory within $\Omega \in [0,96]^{2}$ and the 3D simulation trajectory within $\Omega \in [0,48]^{3}$ in $t \in[0, 750] (s)$, using a $4$th-order Runge–Kutta time integration method~~\cite{ren2022phycrnet} under the periodic condition.
Here, we define $D_u = 0.2, D_v = 0.1, \alpha = 0.025, \beta = 0.055, \Delta t = 0.25 (s)$ and $ \Delta x = 2$.

\paragraph{\textbf{{Black Sea Dataset}.}}
The BS dataset \footnote{ \url{ https://data.marine.copernicus.eu/product/BLKSEA_MULTIYEAR_PHY_007_004/description}} provides measured data of daily mean sea surface temperature $T$ and water flow velocities $\mathbf{u}$ on the Black Sea over several years.
The collection of these data was completed by Euro-Mediterranean Center on Climate Change (CMCC) in Italy starting from June 1, 1993, to June 30, 2021, with a horizontal resolution of  $1/27^{\circ} \times 1/36^{\circ}$.

Note that the first three systems were spatiotemporally down-sampled (e.g., 5-fold in time) from high-fidelity data, while the last one was collected from field measurements.

\subsection{Supplementary details of baseline}
\label{Supplementary Details of Baseline}

\begin{table*}[t]
\caption{{Summary analysis of Baselines.}}
\label{tab:Summary analysis of Baselines}
\vspace{-4pt}
\begin{center}
\begin{small}
\begin{tabular}{lccccccccccc}
\toprule
\textbf{Model} 
& \makecell[c]{\textbf{Grid} \\ \textbf{Domain}} 
& \makecell[c]{\textbf{Irregular} \\ \textbf{Domain}} 
& \makecell[c]{\textbf{Basis function}\\ \textbf{Construction}} 
& \makecell[c]{\textbf{Spectrum}\\ \textbf{Network}}
& \makecell[c]{\textbf{Graph} \\ \textbf{Network}}
& \makecell[c]{\textbf{Attention} \\ \textbf{Network}}
& \makecell[c]{\textbf{Multiscale} \\ \textbf{Modeling}}
& \makecell[c]{\textbf{Geometric} \\ \textbf{Learning}}
& \makecell[c]{\textbf{High-order} \\ \textbf{Modeling}}\\
\midrule
GAT 
&\ding{51}&\ding{51}&\ding{55}&\ding{55}&\ding{51}&\ding{51}&\ding{55}&\ding{55}&\ding{55}\\
\midrule
GATv2 
&\ding{51}&\ding{51}&\ding{55}&\ding{55}&\ding{51}&\ding{51}&\ding{55}&\ding{55}&\ding{55}\\
\midrule
MGN
&\ding{51}&\ding{51}&\ding{55}&\ding{55}&\ding{51}&\ding{55}&\ding{55}&\ding{51}&\ding{55}\\
\midrule
MP-PDE 
&\ding{51}&\ding{51}&\ding{55}&\ding{55}&\ding{51}&\ding{55}&\ding{55}&\ding{51}&\ding{55}\\
\midrule
FNO 
&\ding{51}&\ding{55}&\ding{55}&\ding{51}&\ding{55}&\ding{55}&\ding{55}&\ding{55}&\ding{55}\\
\midrule
FFNO
&\ding{51}&\ding{55}&\ding{51}&\ding{51}&\ding{55}&\ding{55}&\ding{55}&\ding{55}&\ding{55}\\
\midrule
Geo-FNO 
&\ding{51}&\ding{51}&\ding{55}&\ding{51}&\ding{55}&\ding{55}&\ding{55}&\ding{55}&\ding{55}\\
\midrule
Transolver
&\ding{51}&\ding{51}&\ding{55}&\ding{55}&\ding{55}&\ding{51}&\ding{51}&\ding{51}&\ding{55}\\
\midrule
Ours&\ding{51}&\ding{51}&\ding{51}&\ding{55}&\ding{51}&\ding{55}&\ding{55}&\ding{51}&\ding{51}\\
\bottomrule
\end{tabular}
\end{small}
\vspace{-4pt}
\end{center}
\end{table*}


\paragraph{\textbf{{Graph Attention Network (GAT)}.}} 
GAT ~\cite{velickovic2017graph} proposed a graph network with a masked self-attention mechanism, aiming to address the shortcomings of graph convolutions.

\paragraph{\textbf{{Graph Attention Network Variant  (GATv2)}.}}
GATv2 ~\cite{brody2021attentive} proposed a dynamic graph attention variant to remove the limitation of static attention in complex controlled problems, which is strictly more expressive than GAT.





\paragraph{\textbf{{MeshGraphNet~ (MGN)}.}}
MGN ~\cite{pfaff2021Learning} provided a type of neural network architecture designed specifically for modeling physical systems that can be represented as meshes or graphs. Specifically, its `` Encoder-Processor-Decoder'' architecture in ~\cite{pfaff2021Learning} has been widely adopted in many supervised learning tasks, such as fluid and solid mechanics constrained by PDEs. Relevant parameters are referenced from ~\cite{pfaff2021Learning}.

\paragraph{\textbf{{MP-Neural-PDE Solver~ (MP-PDE)}.}}
MP-PDE ~\cite{brandstetter2022message} proposed the temporal bundling and push-forward techniques to encourage zero-stability in training autoregressive models. Relevant parameters are referenced from ~\cite{brandstetter2022message}.

\paragraph{\textbf{{Fourier Neural Operator~ (FNO)}.}}
The most promising spectral approach, FNO ~\cite{li2021Fourier} proposed a neural operator in the Fourier domain to model dynamics. In this work, it was implemented for 2D and 3D spatial grid domains. The hyperparameters are taken from ~\cite{li2021Fourier}.

\paragraph{\textbf{{Factorized Fourier Neural Operator~ (FFNO)}.}}
Factorized Fourier Neural Operator~ (FFNO)~\cite{tran2021factorized} factorizes  the representation into separable Fourier representation to reduce the spatial and temporal complexity, improving its scalability. Relevant parameters are referenced from ~\cite{tran2021factorized}.

\paragraph{\textbf{{Geometry-informed FNO (Geo-FNO)}.}}
Geometry-informed FNO (Geo-FNO) ~\cite{li2023fourier,li2024geometry} maps the irregular domain into a uniform grid, preserving the computation efficiency and handling the arbitrary geometries. Relevant parameters are referenced from ~\cite{li2023fourier}.

\paragraph{\textbf{{Transolver}.}}
Transolver~\cite{wu2024transolver} proposed a new Physics Attention to adaptively split the discretized domain into a series of learnable slices of flexible shapes, effectively capture intricate physical correlations under complex geometrics. Relevant parameters are referenced from ~\cite{wu2024transolver}.

\begin{table*}[t!]
\caption{{Default Training settings for all models. Note that the model's performance does not monotonically improve with increasing parameters. Note that we offer the parameter sets that achieved relative peak performance for each baselines in the prediction tasks.}}
\label{tab:Default Training settings}
\vspace{-4pt}
\begin{center}
\begin{tabular}{lcccccccccc}
\toprule
\textbf{Model} 
& \makecell[c]{\textbf{GAT} } 
& \makecell[c]{\textbf{GATv2}} 
& \makecell[c]{\textbf{MGN}}
& \makecell[c]{\textbf{MP-PDE}}
& \makecell[c]{\textbf{FNO}} 
& \makecell[c]{\textbf{FFNO}}
& \makecell[c]{\textbf{Geo-FNO}}
& \makecell[c]{\textbf{Transolver}}
& \makecell[c]{\textbf{Ours}}\\
\midrule
\makecell[l]{{No. of}  {Layers}}&4&4&4&4&4&4&4&4&4\\
\midrule
\makecell[l]{{Std. of}  {Noise}}&0.0001&0.0001&0.0001&0.0001&0.0001&0.0001&0.0001&0.0001&0.0001\\
\midrule
\makecell[l]{{Learning}  {rate}} &0.0001&0.0001&0.0001&0.0001&0.0001&0.0001&0.0001&0.0001&0.0001\\
\midrule
\makecell[l]{{Hidden}  {dimension}}&128&128&128&128&128&128&128&128&128\\
\midrule
\makecell[l]{{Batch}  {size}} &100&100&100&100&100&100&100&100&100\\
\midrule
\makecell[l]{{No. of}  {Parameters}} &779,014&778,630&514,236&528,658&541,436&1,072,388&1,482,674&1,516,354&1,482,674\\
\bottomrule
\end{tabular}
\vspace{-4pt}
\end{center}
\end{table*}

\begin{table}[htbp]
\caption{Different training hyperparameters for GAT. The default attention head is 1.}
\label{tab:Different hyperparameters for GAT}
\vspace{-4pt}
\begin{center}
\begin{tabular}{lcccccc}
\toprule
\multirow{2}{*}{\makecell[c]{\textbf{Dataset}}}&\multirow{2}{*}{\makecell[c]{\textbf{Attention} \\ \textbf{head}}}&\multirow{2}{*}{\makecell[c]{\textbf{Batch} \\ \textbf{size}}}&\multirow{2}{*}{\makecell[c]{\textbf{No. of} \\ \textbf{Parameters}}}\\
\\
\midrule
\multirow{1}{*}{\makecell[c]{3D GS RD}}   &1      & 5     & 779,014  \\
\multirow{1}{*}{\makecell[c]{2D BS}}      &1    & 20     & 805,350\\
\bottomrule
\end{tabular}
\vspace{-4pt}
\end{center}
\end{table}

\begin{table}[htbp]
\caption{Different training hyperparameters for GATv2. The default attention head is 1.}
\label{tab:Different hyperparameters for GATv2}
\vspace{-4pt}
\begin{center}
\begin{tabular}{lcccccc}
\toprule
\multirow{2}{*}{\makecell[c]{\textbf{Dataset}}}&\multirow{2}{*}{\makecell[c]{\textbf{Attention} \\ \textbf{head}}}&\multirow{2}{*}{\makecell[c]{\textbf{Batch} \\ \textbf{size}}}&\multirow{2}{*}{\makecell[c]{\textbf{No. of} \\ \textbf{Parameters}}}\\
\\
\midrule
\multirow{1}{*}{\makecell[c]{3D GS RD}}   &1   & 5       & 778,630  \\
\multirow{1}{*}{\makecell[c]{2D BS}}      &1   & 20      & 804,966\\
\bottomrule
\end{tabular}
\vspace{-4pt}
\end{center}
\end{table}

\begin{table}[htbp]
\caption{Different training hyperparameters for MGN.}
\label{tab:Different hyperparameters for MGN}
\vspace{-4pt}
\begin{center}
\begin{tabular}{lcccccc}
\toprule
\multirow{2}{*}{\makecell[c]{\textbf{Dataset}}}&\multirow{2}{*}{\makecell[c]{\textbf{Batch} \\ \textbf{size}}}&\multirow{2}{*}{\makecell[c]{\textbf{No. of} \\ \textbf{Parameters}}}\\
\\
\midrule
\multirow{1}{*}{\makecell[c]{3D GS RD}}          & 5     & 514,236  \\
\multirow{1}{*}{\makecell[c]{2D BS}}         & 20     & 540,572  \\
\bottomrule
\end{tabular}
\vspace{-4pt}
\end{center}
\end{table}

\begin{table}[htbp]
\caption{Different training hyperparameters for MP-PDE. The default training length is 2.}
\label{tab:Different hyperparameters for MP-PDE}
\vspace{-4pt}
\begin{center}
\begin{tabular}{lcccccc}
\toprule
\multirow{2}{*}{\makecell[c]{\textbf{Dataset}}}&\multirow{2}{*}{\makecell[c]{\textbf{Training} \\ \textbf{length}}}&\multirow{2}{*}{\makecell[c]{\textbf{Batch} \\ \textbf{size}}}&\multirow{2}{*}{\makecell[c]{\textbf{No. of} \\ \textbf{Parameters}}}\\
\\
\midrule
\multirow{1}{*}{\makecell[c]{3D GS RD}}   &2    & 5     & 528,658  \\
\multirow{1}{*}{\makecell[c]{2D BS}}      &2  & 20     & 554,994  \\
\bottomrule
\end{tabular}
\vspace{-4pt}
\end{center}
\end{table}

\begin{table}[htbp]
\caption{Different training hyperparameters for FNO. The default mode size is 8.}
\label{tab:Different hyperparameters for FNO}
\vspace{-4pt}
\begin{center}
\begin{tabular}{lcccccc}
\toprule
\multirow{2}{*}{\makecell[c]{\textbf{Dataset}}}&\multirow{2}{*}{\makecell[c]{\textbf{No. of} \\ \textbf{mode}}}&\multirow{2}{*}{\makecell[c]{\textbf{Batch} \\ \textbf{size}}}&\multirow{2}{*}{\makecell[c]{\textbf{No. of} \\ \textbf{Parameters}}}\\
\\
\midrule
\multirow{1}{*}{\makecell[c]{3D GS RD}}         & 8       & 10     & 4,515,746  \\
\bottomrule
\end{tabular}
\vspace{-4pt}
\end{center}
\end{table}

\begin{table}[htbp]
\caption{Different training hyperparameters for FFNO. The default mode size is 8.}
\label{tab:Different hyperparameters for FFNO}
\vspace{-4pt}
\begin{center}
\begin{tabular}{lcccccc}
\toprule
\multirow{2}{*}{\makecell[c]{\textbf{Dataset}}}&\multirow{2}{*}{\makecell[c]{\textbf{No. of} \\ \textbf{mode}}}&\multirow{2}{*}{\makecell[c]{\textbf{Batch} \\ \textbf{size}}}&\multirow{2}{*}{\makecell[c]{\textbf{No. of} \\ \textbf{Parameters}}}\\
\\
\midrule
\multirow{1}{*}{\makecell[c]{3D GS RD}}   &8       & 10     &  1,334,660 \\
\bottomrule
\end{tabular}
\vspace{-4pt}
\end{center}
\end{table}

\begin{table}[htbp]
\caption{Different training hyperparameters for Geo-FNO. The default mode size is 8 and the mapping size is 32$\times$32.}
\label{tab:Different hyperparameters for Geo-FNO}
\vspace{-4pt}
\begin{center}
\begin{tabular}{lcccccc}
\toprule
\multirow{2}{*}{\makecell[c]{\textbf{Dataset}}}&\multirow{2}{*}{\makecell[c]{\textbf{No. of} \\ \textbf{mode}}}&\multirow{2}{*}{\makecell[c]{\textbf{Mapping} \\ \textbf{size}}}&\multirow{2}{*}{\makecell[c]{\textbf{Batch} \\ \textbf{size}}}&\multirow{2}{*}{\makecell[c]{\textbf{No. of} \\ \textbf{Parameters}}}\\
\\
\midrule
\multirow{1}{*}{\makecell[c]{3D GS RD}}  &8 &32 $\times$ 32 $\times$ 32  & 10     &  2,705,569 \\
\multirow{1}{*}{\makecell[c]{2D BS}}     &8 &32 $\times$ 32  & 20     & 727,557  \\
\bottomrule
\end{tabular}
\vspace{-4pt}
\end{center}
\end{table}

\begin{table}[htbp]
\caption{Different training hyperparameters for Transolver. The default attention head is 8 and the mapping size is 8$\times$8.}
\label{tab:Different hyperparameters for Transolver}
\vspace{-4pt}
\begin{center}
\begin{tabular}{lcccccc}
\toprule
\multirow{2}{*}{\makecell[c]{\textbf{Dataset}}}&\multirow{2}{*}{\makecell[c]{\textbf{Attention} \\ \textbf{head}}}&\multirow{2}{*}{\makecell[c]{\textbf{Mapping} \\ \textbf{size}}}&\multirow{2}{*}{\makecell[c]{\textbf{Batch} \\ \textbf{size}}}&\multirow{2}{*}{\makecell[c]{\textbf{No. of} \\ \textbf{Parameters}}}\\
\\
\midrule
\multirow{1}{*}{\makecell[c]{3D GS RD}}  &8  &8 $\times$ 8 $\times$8   & 10     &  1,631,042 \\
\multirow{1}{*}{\makecell[c]{2D BS}}    &8  &8 $\times$ 8  & 20     &  1,516,739 \\
\bottomrule
\end{tabular}
\vspace{-4pt}
\end{center}
\end{table}

\begin{table}[htbp]
\caption{Different training hyperparameters for CeFeGNN. The default window size is 4.}
\label{tab:Different hyperparameters for CeFeGNN}
\vspace{-4pt}
\begin{center}
\begin{tabular}{lcccccc}
\toprule
\multirow{2}{*}{\makecell[c]{\textbf{Dataset}}}&\multirow{2}{*}{\makecell[c]{\textbf{Window} \\ \textbf{size}}}&\multirow{2}{*}{\makecell[c]{\textbf{Batch} \\ \textbf{size}}}&\multirow{2}{*}{\makecell[c]{\textbf{No. of} \\ \textbf{Parameters}}}\\
\\
\midrule
\multirow{1}{*}{\makecell[c]{3D GS RD}}   &4  & 5     & 1,482,674  \\
\multirow{1}{*}{\makecell[c]{2D BS}}      &4  & 20     & 1,509,010  \\
\bottomrule
\end{tabular}
\vspace{-4pt}
\end{center}
\end{table}


\subsection{Supplementary details of results}
\label{Supplementary Details of results}

\subsubsection{Summary of parameters of all models}

\paragraph{\textbf{Overall settings}} As shown in Table \ref{tab:Default Training settings}, the default number of layers is 4; the learning rate is $1 \times 10^{-4}$; the hidden dimension is 128; the batch size is 100; the standard deviation of training noise is $1 \times 10^{-4}$. We offer the basic parameter sets of each baselines in the prediction tasks.

In addition, we list the different parameters of all models on various datasets in Tables \ref{tab:Different hyperparameters for GAT} to \ref{tab:Different hyperparameters for CeFeGNN}, such as the mode size in FNO and the number of head in the attention mechanism.

\subsubsection{Data scaling test}

We also have performed a data scaling test and report the results (data size vs. prediction error) in the Table \ref{tab:data_scaling_results} below to support our claim of low data requirement.

\begin{table}[htbp]
\caption{Data scaling test under RMSE metric of CeFeGNN, MGN and MP-PDE on 2D Burgers equation.}
\label{tab:data_scaling_results}
\vspace{-4pt}
\begin{center}
\begin{small}
\begin{tabular}{lcc|ccc}
\toprule
\multirow{2}{*}{\textbf{Model}}
&\multirow{2}{*}{\makecell[c]{\textbf{No.of} \\ \textbf{layers}}}
&\multirow{2}{*}{\makecell[c]{\textbf{Dimension}}}
&\multicolumn{3}{c}{\textbf{Data size }}\\
\cmidrule{4-6}
&&
& \makecell[c]{{30}} 
& \makecell[c]{{40}} 
& \makecell[c]{{50}} \\
\midrule
CeFeGNN &4   &128 &\textbf{0.00664}&\textbf{0.00413}& \textbf{0.00226}\\
MGN   &4   &128 &\underline{0.01174}&\underline{0.00926}&\underline{0.00636}\\
MP-PDE&4   &128 &0.01784&0.01442&0.00911\\
\bottomrule
\end{tabular}
\end{small}
\end{center}
\vspace{-4pt}
\end{table}

\subsubsection{Discussion of the position information in cells}

In this part, we provide other two cases ``CeFeGNN w/o Cell Pos.'' and `` w Pos. to B'' in the following Table \ref{tab:cell_pos}, where `` w/o Cell Pos.'' represents the cell initial feature without the position awareness, and `` w Pos. to B'' is replacing the distance to the cell center with the distance to the nearest PDE boundary. The results in Table \ref{tab:cell_pos} show that the performance improvement of CeFeGNN is not only due to add position awareness into the cell initial feature, but also the second-order refinement of the discrete space. 

\begin{table}[htbp]
\caption{Quantitative ablation results about the cell position information for CeFeGNN and other two cases under RMSE metrics.
}
\label{tab:cell_pos}
\vspace{-4pt}
\begin{center}
\begin{small}
\begin{tabular}{l|cccc|c}
\toprule
\multirow{4}{*}{\textbf{Model}}&\multicolumn{5}{c}{\textbf{RMSE $\downarrow$}}\\
\cmidrule{2-6}
& \makecell[c]{\textbf{ Burgers} \\ ($0.001 s$)} 
& \makecell[c]{\textbf{ FN}  \\ ($0.002 s$)} 
& \makecell[c]{\textbf{2D GS }  \\ ($0.25 s$)} 
& \makecell[c]{\textbf{3D GS }  \\ ($0.25 s$)}
& \makecell[c]{\textbf{ BS}  \\ ($1 \text{day}$)}\\
\midrule
CeFeGNN &\textbf{0.00664}&\textbf{0.00364}&\textbf{0.00248}&\textbf{0.00138}&\textbf{0.55599}\\
\textbf{w/o} C Pos. &0.00721&\underline{0.00477}&\underline{0.00439}&\underline{0.00274}&0.56098\\
\textbf{w} Pos. to B &\underline{0.00720}&0.00490&0.00445&0.00276&\underline{0.56040}\\
MGN &0.01174&0.02108&0.02917&0.01925&0.61475\\
\bottomrule
\end{tabular}
\end{small}
\end{center}
\vspace{-4pt}
\end{table}

\subsubsection{Discussion with other Higher-order Graphs}

In this part, we mainly focus on discussing the excellent works CCNNs in ~\cite{hajij2022topological} and MPSNs in ~\cite{bodnar2021weisfeiler}. A summary of these models is shown in the following Table \ref{tab:discuss_higher_order_graphs}. 

\begin{table}[htbp]
\caption{Summary of CeFeGNN, MGN, CCNNs, and MPSNs}
\label{tab:discuss_higher_order_graphs}
\vspace{-4pt}
\begin{center}
\begin{small}
\begin{tabular}{lcccc}
\toprule
\multirow{2}{*}{\textbf{Model}}&\multirow{2}{*}{\makecell[c]{\textbf{Message} \\ \textbf{Level}}}&\multirow{2}{*}{\makecell[c]{\textbf{Complex} \\ \textbf{Predefinition}}}&\multirow{2}{*}{\makecell[c]{\textbf{Complex} \\ \textbf{Type}}}&\multirow{2}{*}{\textbf{Application}}\\
\\
\midrule
CeFeGNN    & 2 & No & 3 & Dynamics \\
MGN      & 2 & No & 2 & Dynamics \\
CCNNs & 3 & No & 3 & Classification \\
MPSNs & 2 & Yes & 5 & Classification \\
\bottomrule
\end{tabular}
\end{small}
\end{center}
\vspace{-4pt}
\end{table}

Firstly, MGN achieves message passing through a two-level structure ($edge \rightarrow node$). On this basis, CCNNs in ~\cite{hajij2022topological} leverages combinatorial complexes to achieve message passing through a three-level structure ($cell \rightarrow edge \rightarrow node$). 
Meanwhile, MPSNs in ~\cite{bodnar2021weisfeiler} performs message passing through a two-level structure ($complexes \rightarrow simplex$) on various simplicial complexes (SCs), which primarily include one simplex (node) and four types of complexes with varying adjacent simplices (e.g., boundary adjacencies, co-boundary adjacencies, lower-adjacencies, and upper-adjacencies), to enhance feature distinguishability and thus improve classification performance.

In contrast, CeFeGNN sets apart from these works and adopts a novel two-level structure ($[cell, edge] \rightarrow node$) from the spatial perspective, which is more suitable to learn implicit dynamic mechanism.

Since there are only node labels for the supervised learning, the three-level message passing mechanism in ~\cite{hajij2022topological} poses significant training challenges. However, our two-level message passing sequence reduces the high coupling degree in ~\cite{hajij2022topological} and avoids the limitation for additional predefined special complexes (e.g., some simplicial complexes in ~\cite{bodnar2021weisfeiler} ). A comparison result between two-level and three-level mechanisms is shown in the following Table \ref{tab:message_level}. The results in Table \ref{tab:message_level} demonstrate that three-level mechanism underperforms than two-level mechanism in our all cases. 

\begin{table}[htbp]
\caption{Comparison result between two-level and three-level mechanisms with various time intervals ($\Delta t$).
}
\label{tab:message_level}
\vspace{-4pt}
\begin{center}
\begin{small}
\begin{tabular}{l|cccc|c}
\toprule
\multirow{4}{*}{\textbf{Model}}&\multicolumn{5}{c}{\textbf{RMSE $\downarrow$}}\\
\cmidrule{2-6}
& \makecell[c]{\textbf{ Burgers} \\ ($ 0.001 s$)} 
& \makecell[c]{\textbf{ FN}  \\ ($0.002 s$)} 
& \makecell[c]{\textbf{2D GS }  \\ ($ 0.25 s$)} 
& \makecell[c]{\textbf{3D GS }  \\ ($0.25 s$)}
& \makecell[c]{\textbf{ BS}  \\ ($ 1 \text{day}$)}\\
\midrule
CeFeGNN &\textbf{0.00664}&\textbf{0.00364}&\textbf{0.00248}&\textbf{0.00138}&\textbf{0.55599}\\
MGN (2-level) &\underline{0.01174}&\underline{0.02108}&\underline{0.02917}&\underline{0.01925}&\underline{0.61475}\\
MGN (3-level) &0.03220&0.04196&0.05884&0.08218&0.61524\\
\bottomrule
\end{tabular}
\end{small}
\end{center}
\vspace{-4pt}
\end{table}

\begin{table}[htbp]
\caption{Statistic information of traffic datasets (PeMS03 and PeMS04). Here, ``F'' represents the traffic flow, ``S'' the traffic speed, and ``O'' the traffic occupancy rate.}
\label{tab:traffic_dataset}
\vspace{-4pt}
\begin{center}
\begin{small}
\begin{tabular}{l|cc}
\toprule
\multirow{1}{*}{\textbf{Property}}&\multicolumn{1}{c}{\textbf{PeMS03}}&\multicolumn{1}{c}{\textbf{PeMS04}}\\
\midrule
No. of nodes &358	&307\\
No. of time steps &26,208	&16,992\\
Sampling time	&5 min	& 5 min\\
Time range (in 2018)	&09.01--11.30	&01.01--02.28\\
Variable	&F	&F, S, O\\
Training/Validation/Test &0.6/0.2/0.2  &0.6/0.2/0.2 \\
\bottomrule
\end{tabular}
\end{small}
\end{center}
\vspace{-4pt}
\end{table}

\begin{table}[htbp]
\caption{Ablation Results on PeMS03 and PeMS04 with the MAE, RMSE, and MAPE ($\%$) metric. The abbreviation “C” represents the cell.}
\label{tab:ablation_results_on_traffic}
\vspace{-4pt}
\begin{center}
\begin{small}
\begin{tabular}{l|ccc|ccc}
\toprule
\multirow{2}{*}{\textbf{Model}}&\multicolumn{3}{c}{\textbf{PeMS03 $\downarrow$}}&\multicolumn{3}{c}{\textbf{PeMS04 $\downarrow$}}\\
\cmidrule{2-7}
&MAE&RMSE&MAPE&MAE &RMSE&MAPE\\
\midrule
CeFeGNN &	\textbf{14.08}&\textbf{24.14}&\textbf{14.45}&	\textbf{17.91}&\textbf{29.28}&\textbf{11.86}\\
\textbf{w/o} C&	\underline{14.87}&25.66&\underline{14.95}&	\underline{18.21}&\underline{29.82}&\underline{11.98}\\
\textbf{w/o} FE& 14.92&\underline{25.42}&14.99&	18.29&29.91&11.99\\
\textbf{w/o} C, FE&	15.31&27.33&15.14&	18.40&30.26&12.13\\
\bottomrule
\end{tabular}
\end{small}
\end{center}
\vspace{-4pt}
\end{table}

\subsubsection{Extension on traffic prediction}\label{Extension on traffic prediction}
For providing a more comprehensive evaluation, we conduct experimnents on traffic forecasting tasks, where stronger empirical evidence could better substantiate the model's effectiveness. As shown in Table \ref{tab:results_on_traffic}, we provide the results on traffic datasets (PeMS03 and PeMS04, \footnote{\url{ https://pems.dot.ca.gov/}} see Table \ref{tab:traffic_dataset} for more details.) and consider more related baselines, including StGCN \cite{yu2017spatio}, ASTGCN \cite{zhao2022attention}, AGCRN \cite{bai2020adaptive}, STSGCN \cite{song2020spatial}, STFGNN \cite{li2021spatial}, ASTGNN \cite{zhou2021ast}, PDFormer \cite{jiang2023pdformer}, STAEformer \cite{liu2023staeformer}, HTVGNN \cite{dai2024novel}, FasterSTS \cite{dai2025fastersts}, and DTRformer \cite{chen2025dynamic}. We can see that our model achieved overall the best performance.


In addition, a ablation study on PeMS03 and PeMS04 is provided in Table \ref{tab:ablation_results_on_traffic} to show the contributions of each components in our model. These results demonstrate the effectiveness of CeFeGNN.

\subsubsection{Physical loss of PDE systems}
In this work, our research objective is to learn a general model that can be applied in various scenarios, such as real-world prediction tasks (ocean data prediction and traffic prediction). Therefore, we \textbf{did not} incorporate hard physical constraints in the \textbf{training} process.

To evaluate the physical consistency of the model prediction on PDE systems, we added an analysis of the PDE loss during \textbf{inference (not training)} of our model, with the results summarized in Table \ref{tab:results_of_physical_loss}. This allows us to quantitatively assess our model's adherence to underlying physical principles without compromising its data-driven nature. It can be seen from Table \ref{tab:results_of_physical_loss} that our model achieved distinctively higher physical consistency compared with other suboptimal models (MGN, MP-PDE) in the paper.

\begin{remark}
Our work learns spatiotemporal mappings directly from data (\textbf{data-driven}) applicable to generic systems \textbf{without prior knowledge} (e.g., a strong explicit prior knowledge in \textbf{equation-driven} PINNs) because PINNs fail to handle the complicated real-world BS example since \textbf{no convincing governing equations are available}.
\end{remark}

\begin{table}[htbp]
\caption{Results on PeMS03 and PeMS04 with the MAE, RMSE, and MAPE ($\%$) metric. }
\label{tab:results_on_traffic}
\vspace{-4pt}
\begin{center}
\begin{small}
\begin{tabular}{l|ccc|ccc}
\toprule
\multirow{2}{*}{\textbf{Model}}&\multicolumn{3}{c}{\textbf{PeMS03 $\downarrow$}}&\multicolumn{3}{c}{\textbf{PeMS04 $\downarrow$}}\\
\cmidrule{2-7}
&MAE&RMSE&MAPE&MAE &RMSE&MAPE\\
\midrule
Transformer &	17.50&30.24&16.80&	23.83&37.19&15.57\\
STGCN &	17.49&30.12&17.15&	22.70&35.55&14.59\\
ASTGCN &	17.69&29.66&19.40&	22.93&35.22&16.56\\
AGCRN&	16.06&28.49&15.85&	19.83&32.26&12.97\\
STSGCN &	17.48&29.21&16.78&	21.19&33.65&13.90\\
STFGNN &	16.77&28.34&16.30&	19.83&31.88&13.02\\
ASTGNN &	15.07&26.88&15.80&	19.26&31.16&12.65\\
MGN &15.31 &27.33 &15.14 &18.40 &30.26 &12.13 \\
PDFormer &	14.94&25.39&15.82&	18.32&29.97&12.10\\
STAEformer &	15.35&27.55&15.18&	18.22&30.18&11.98\\
HTVGNN &	\underline{14.20}&24.26&\underline{14.52}&	\underline{17.99}&\underline{29.74}&\underline{11.90}\\
FasterSTS &	14.58&\textbf{24.12}&14.93&	18.49&29.92&12.21\\
DTRformer &	14.50&25.45&14.94&	18.00&29.58&12.30\\
CeFeGNN&	\textbf{14.08}&\underline{24.14}&\textbf{14.45}&\textbf{17.91}&\textbf{29.28}&\textbf{11.86}\\
\bottomrule
\end{tabular}
\end{small}
\end{center}
\vspace{-4pt}
\end{table}

\begin{table}[htbp]
\caption{Results of physical loss (aka, PDE residual loss).}
\label{tab:results_of_physical_loss}
\vspace{-4pt}
\begin{center}
\begin{small}
\begin{tabular}{l|ccc}
\toprule
\multirow{1}{*}{\textbf{Model}}&\multicolumn{1}{c}{\textbf{2D Burgers $\downarrow$}}&\multicolumn{1}{c}{\textbf{2D FN $\downarrow$}}&\multicolumn{1}{c}{\textbf{2D GS $\downarrow$}}\\
\midrule
CeFeGNN &	\textbf{6.412} $\times 10^{-5}$&\textbf{6.994} $\times 10^{-5}$&\textbf{2.854} $\times 10^{-5}$\\
MGN&	\underline{12.01} $\times 10^{-5}$&\underline{38.39} $\times 10^{-5}$&\underline{45.11} $\times 10^{-5}$\\
MP-PDE&	18.96 $\times 10^{-5}$&67.05 $\times 10^{-5}$&87.53 $\times 10^{-5}$\\
\bottomrule
\end{tabular}
\end{small}
\end{center}
\vspace{-4pt}
\end{table}

\subsubsection{Effect of several temporal models on CeFeGNN}
In our work, we employed \textbf{residual connections to model temporal marching} (see Eq. \ref{eq:decoder}). Our model architecture can be summarized as "encoder $\rightarrow$ processor $\rightarrow$ decoder". Here we insert the TAN or LSTM before "decoder" (i.e., encoder $\rightarrow$ processor $\rightarrow$ TAN/LSTM $\rightarrow$ decoder). Following P0, we firstly set the training input step as 1 (same as the setting of CeGNN) with the results in Table \ref{tab:results_of_varying_k}.

Obviously, the reason for the TAN and LSTM failing in the case of 1 input step is that they do not learn the temporal dependencies as expected, but disrupt the spatiotemporal information transmission of the basic model.

\begin{figure}[htbp]
	\centering
	\vspace{-4pt}
	\includegraphics[width=0.8\columnwidth]{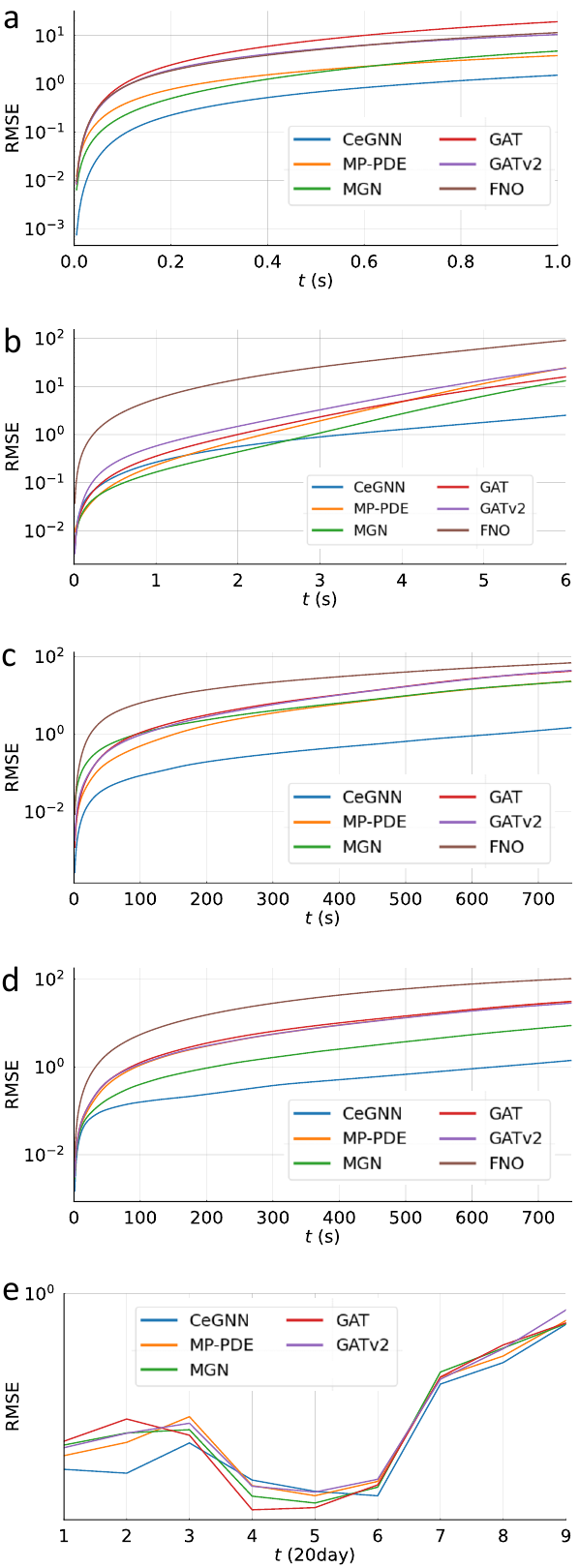}
	\caption{Error propagation curve on various systems. \textbf{a}, 2D Burgers equation. \textbf{b}, 2D FN equation. \textbf{c}, 2D GS equation. \textbf{d}, 3D GS equation. \textbf{e}, 2D BS dataset.}
	\label{fig:error_curve}
	\vspace{-4pt}
\end{figure}

Therefore, we further conducted experiments with the setting of "N $\rightarrow$ N" (aka, next token generation; herein, we took N = 3 for illustration purpose) and summarized the results in Table \ref{tab:results_of_varying_k}. We can see that the temporal methods (TAN, LSTM) yield a performance improvement in such a setting (however, still inferior to our model). This might be because such temporal modules add extra model complexity, which might bring over-fitting issues especially when the training data is limited. Note that the attention mechanism and the LSTM module bring a 2 $\times$ increase of training speed.

\begin{table}[htbp]
\caption{Results of varying training input length (k) on 2D Burgers equation.}
\label{tab:results_of_varying_k}
\vspace{-4pt}
\begin{center}
\begin{small}
\begin{tabular}{l|c|cc}
\toprule
\multirow{3}{*}{\textbf{Input step}}&\multicolumn{1}{c}{\textbf{k=1 }}&\multicolumn{2}{c}{\textbf{k=3}}\\
\cmidrule{2-4}
&RMSE $\downarrow$ & RMSE $\downarrow$&\makecell[c]{Training time \\ (s/epoch) }\\
\midrule
CeFeGNN &	\textbf{0.0066} &\textbf{0.0049} &\textbf{17.31} \\
CeFeGNN + TAN&	\underline{0.0981} &\underline{0.0584} &\underline{38.13} \\
CeFeGNN + LSTM&	0.1293 &0.0689 &39.86 \\
\bottomrule
\end{tabular}
\end{small}
\end{center}
\vspace{-4pt}
\end{table}

\subsubsection{Error propagation curve}\label{Error propagation curve}
In this part, we provide the error propagation curves of CeFeGNN and several representative models on all benchmarks, as shown in Figure \ref{fig:error_curve}.

\subsubsection{Full snapshots}\label{Full snapshots}
\begin{figure*}[htbp]
\centering
\vspace{-4pt}
\includegraphics[width=\textwidth]{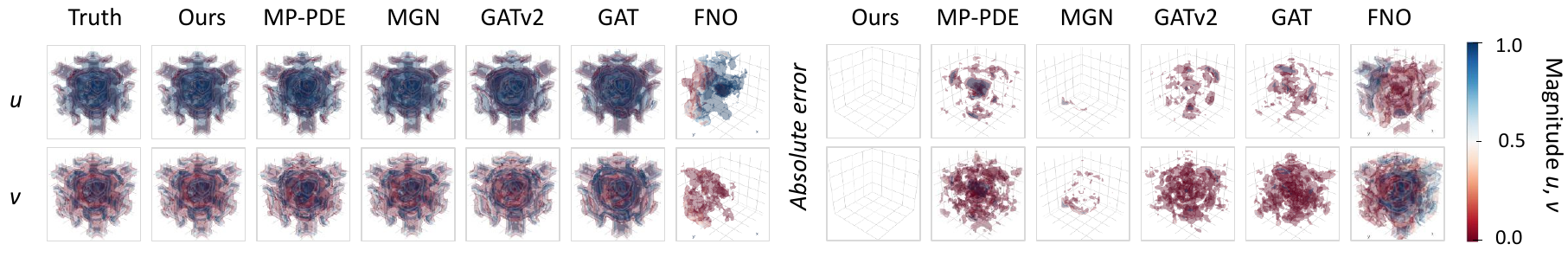}
\caption{Full snapshots on 3D GS equation. The absolute error represents the absolute difference between ground-truth data and the prediction values.}
\label{fig:snapshot_3dgs}
\vspace{-4pt}
\end{figure*}

\begin{figure*}[htbp]
\centering
\vspace{-4pt}
\includegraphics[width=\textwidth]{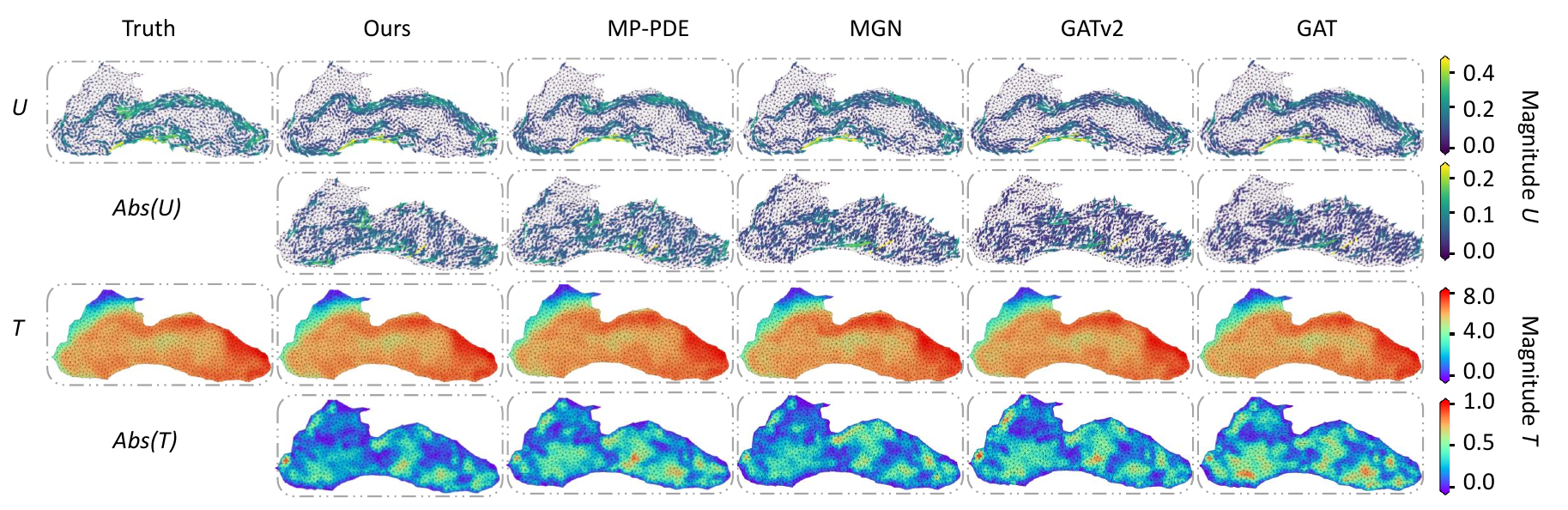}
\caption{Full snapshots on 2D BS dataset. The ``Abs($\cdot$)'' represents the absolute difference between ground-truth data and the prediction values}
\label{fig:snapshot_2dbs}
\vspace{-4pt}
\end{figure*}

In this part, we provide the full snapshots of CeFeGNN and several representative models on 3D GS and 2D BS benchmarks, as shown in Figures \ref{fig:snapshot_3dgs} and \ref{fig:snapshot_2dbs}.

\clearpage

\end{document}